\documentclass[letterpaper]{article} 
\usepackage{aaai25}  
\usepackage{times}  
\usepackage{helvet}  
\usepackage{courier}  
\usepackage[hyphens]{url}  
\usepackage{graphicx} 
\usepackage{natbib}  
\usepackage{caption} 
\usepackage{algorithm}
\usepackage{algorithmic}
\usepackage{amsmath}
\usepackage{multirow}
\usepackage{newfloat}
\usepackage{listings}
\DeclareCaptionStyle{ruled}{labelfont=normalfont,labelsep=colon,strut=off} 
\floatstyle{ruled}
\newfloat{listing}{tb}{lst}{}
\floatname{listing}{Listing}
\usepackage{booktabs,subcaption,amsfonts,dcolumn}

\usepackage[capitalise,nameinlink]{cleveref}
\crefname{section}{Sec.}{Secs.}
\crefname{algorithm}{Alg.}{Algs.}
\crefname{appendix}{App.}{Apps.}
\crefname{definition}{Def.}{Defs.}
\crefname{table}{Table}{Tables}
\usepackage{cleveref}
\usepackage{amssymb}
\usepackage{todonotes}
\usepackage{xspace}
\newcommand{\eg}{\textit{e.g.\@}\xspace}
\newcommand{\ie}{\textit{i.e.\@}\xspace}

\newcommand{\our}{\textsc{er-trl}\xspace}

\title{Entropy Regularized Task Representation Learning for Offline Meta-Reinforcement Learning}
\author {
    Mohammadreza Nakhaeinezhadfard\textsuperscript{\rm 1},
    Aidan Scannell\textsuperscript{\rm 1,\rm 2},
    Joni Pajarinen\textsuperscript{\rm 1}
}
\affiliations {
    \textsuperscript{\rm 1}Department of Electrical Engineering and Automation (EEA), Aalto University, Finland\\
    \textsuperscript{\rm 2}Finnish Center for Artificial Intelligence, Finland\\
    \{mohammadreza.nakhaei, aidan.scannell, joni.pajarinen\}@aalto.fi 
}
\begin{document}
\maketitle
\begin{abstract}
Offline meta-reinforcement learning aims to equip agents with the ability to rapidly adapt to new tasks by training on data from a set of different tasks.
Context-based approaches utilize a history of state-action-reward transitions -- referred to as the context -- to infer representations of the current task, and then condition the agent, i.e., the policy and value function, on the task representations.
Intuitively, the better the task representations capture the underlying tasks, the better the agent can generalize to new tasks.
Unfortunately, context-based approaches suffer from distribution mismatch, as the context in the offline data does not match the context at test time, limiting their ability to generalize to the test tasks.
This leads to the task representations overfitting to the offline training data.
Intuitively, the task representations should be independent of the behavior policy used to collect the offline data.
To address this issue, we approximately minimize the mutual information between the distribution over the task representations and behavior policy by maximizing the entropy of behavior policy conditioned on the task representations. 
We validate our approach in MuJoCo environments, showing that compared to baselines, our task representations more faithfully represent the underlying tasks, leading to outperforming prior methods in both in-distribution and out-of-distribution tasks.
\end{abstract}

%
\begin{links}
    \link{Code}{https://github.com/MohammadrezaNakhaei/ER-TRL}
\end{links}
\section{Introduction}
The goal of offline reinforcement learning (RL; \citealp{survey, levine_tutorial}) is to leverage offline datasets to learn policies that can accomplish tasks without interacting with the environment. 
It is a promising approach for real-world applications where online RL may be expensive or dangerous, such as robotics \citep{offlinerl_realworld, offlinerl_robotic_skill} and healthcare \citep{offlinerl_blood, offlinerl_insulin}. 
Offline meta-RL (OMRL) extends offline RL to the meta-RL setting, where the goal is to solve a previously unseen task, by leveraging behaviors learned from a set of training environments. 
In contrast to meta-RL, OMRL considers the setting where the agent does not directly interact with the training environments but instead has access to offline datasets from each training environment.

A promising approach to OMRL is context encoding \citep{mbml, focal, corro, context_distribution_shift, idac, csro}, where a context encoder learns a representation of the task from a history of state-action-reward transitions, referred to as context. 
The agent is then conditioned on this learned task representation. 
For example, in off-policy RL, the policy and value functions are conditioned on the task representation, which
enables adaptation to different tasks, including previously unseen ones.

As highlighted by \citet{csro}, context-based OMRL suffers from a distribution shift in the context encoder.
This is because the context encoder is trained with contexts
collected by the behavior policy, 
but at test time the context is collected by a different policy, including the learned policy conditioned on a prior task representation. 
As such, there is a discrepancy between the contexts at train and test time. 
This limits the context encoder's ability to infer the correct task at test time, 
which consequently limits the agent's ability to adapt to previously unseen tasks.

One approach to alleviating this distribution mismatch is to minimize the mutual information between the task representations and the behavior policy. 
The intuition is that for environments that require dissimilar policies for each task, the context encoder should not rely on the characteristics of the behavior policy.
As such, we would like our learned task representations to be independent of the behavior policy in order to reduce the impact of the behavior policy on the task representation.
We sidestep the context shift issue by showing that minimizing the mutual information between the task representations and the behavior policy is equivalent to maximizing the entropy of a meta-behavior policy conditioned on the learned task representation.

\textbf{Our contributions} are as follows:
\begin{itemize}
    \item We present \textit{Entropy Regularized Task Representation Learning} (\our), an OMRL method that improves the ability of context-based OMRL methods to generalize to previously unseen environments. \our reduces the amount of context distribution shift by leveraging a generative adversarial network (GAN; \citealp{gan}) to minimize the mutual information between the task representations and the behavior policy, by estimating the entropy of the meta-behavior policy.
    \item We show that our task representation learning improves performance in in-distribution tasks and also enhances generalization to out-of-distribution tasks.
    \item We then show that \our's task representation learning outperforms prior methods as it can better predict the true task representation, \eg the target velocity or direction in locomotion tasks. 
\end{itemize}

\section{Background}
This section formally defines the context-based OMRL framework and subsequently explores the critical challenge of context shift, where the distribution of contexts during training and testing diverges, significantly impacting the performance of context-based OMRL algorithms.

\paragraph{Context-based Offline Meta-RL}
In OMRL, there is a distribution over tasks where each task is represented as a Markov Decision Process (MDP), ${\mathcal{M}_i = \langle \mathcal{S}, \mathcal{A}, R_i, P_i, \gamma, \rho_0 \rangle}$, consisting of a shared state space $\mathcal{S}$, action space $\mathcal{A}$, discount factor $\gamma \in [0,1]$, and initial state distribution $\rho_0(s_0)$, and task-specific reward function $R_i: \mathcal{S} \times \mathcal{A} \rightarrow \mathbb{R}$ and transition dynamics $P_i (s_{t+1}|s_t, a_t)$, which represents the distribution of possible next states conditioned on the current state and action.  
The objective is to train a meta-policy $\pi$ that can generalize to new tasks, \ie, maximizing the expected cumulative reward over the distribution of tasks 
\begin{align} \label{eq:objective}
    J(\pi) = \mathbb{E}_{\substack{\mathcal{M}_i \sim p(\mathcal{M}) \\ s_0 \sim \rho_0(s_0) \\ s_{t+1} \sim P_i(\cdot \mid s_t, a_t) \\ a_t \sim \pi(a_t \mid \cdot)}} 
    \left [\sum_{t=0}^T \gamma^t R_i (s_t,a_t) \right].
\end{align}
A common assumption for OMRL is that the reward function and the transition dynamics for all tasks are deterministic. 

The offline dataset $\mathcal{D}_i = \{(s_{j}, a_{j}, r_{j}, s_{j}^\prime)^i\}_{j=1}^{n_i} $ for task $\mathcal{M}_i$ consists of state, action, reward, next state tuples collected using its corresponding behavior policy $\pi_\beta^i (a|s)$. 
During training, context-based OMRL methods use a mini-batch of transitions from the offline dataset $c_i \subset \mathcal{D}_i$ -- referred to as the context -- to infer the task. 
A context encoder $e_{\theta}$ learns a mapping from the context $c_i$ to latent task representations $z_i$.
The policy and value functions are then conditioned on the task representations $z_i$ and trained to maximize \cref{eq:objective} using the offline datasets. Note that there is no online interaction. 

\paragraph{Distribution Shift in Context Encoder}
During training, the context encoder is trained on offline datasets, collected by the behavior policy. 
The distribution over offline contexts $p(c_i)$ depends on the transition dynamics, reward function, and the behavior policy:
\begin{equation*}
p(c_i) = p(s_1) \prod_{j=1}^n \pi_\beta^i (a_j|s_j) R_i(s_j, a_j) P_i(s_{j+1}|s_j ,a_j).  
\end{equation*}
However, at test time, the context is collected with an exploration policy $\pi_{\text{explore}}$, which may be {\em (i)} the learned policy conditioned on some prior task representations $z_0$, {\em (ii)} a random policy, or {\em (iii)} a combination of them. 
As such, the difference in the behavior and exploration policies leads to the distribution over the context being different at train and test time
\begin{equation*}
    p(c_i^\text{test}) = p(s_1) \prod_{j=1}^n \pi_{\text{explore}} (a_j|.) R_i(s_j, a_j) P_i(s_{j+1}|s_j ,a_j).  
\end{equation*}
Ideally, the context encoder should compress task-related information in the context and since the transition dynamics and reward function are different across tasks, the learned representation should only depend on them. 
However, as illustrated, the behavior policy affects the context distribution so the context encoder embeds some characteristics of the behavior policy specific to the task. 
In testing, the exploration policy's characteristics might differ, leading to inferring an irrelevant task representations and causing a reduction in performance. 

\section{Related Work}
In this section, we provide an overview of context-based OMRL methods. We aim to highlight how different approaches attempt to tackle the distribution shift problem, which in turn motivates our approach.

\paragraph{Pre-collected contexts}
Some methods \citep{mbml, focal, corro} sidestep the issue of distribution shift in the context by collecting datasets from the test tasks using the behavior policy and using these to infer the task representations. 
Whilst these approaches sidestep the distribution shift problem, we highlight that this setting is unrealistic since it requires the agent to have access to all of the test tasks a priori, such that contexts for the test tasks can be collected with the behavior policy.
In contrast, we consider the more realistic setting, where the agent does not have access to the test tasks a priori. 
It collects the context by online interaction with the environment. 
\paragraph{Context filtering}
IDAC \cite{idac} addresses the context shift problem by filtering transitions while collecting the context. 
They utilize uncertainty quantification, \eg prediction error of an ensemble of dynamics models, to filter out-of-distribution trajectories in the context. 
Trajectories with higher uncertainty than a pre-defined threshold are disregarded in the context.
Therefore, the collected context contains in-distribution transitions for inferring the task. 
However, this approach requires more interaction with the environment and is brittle to the choice of the threshold for uncertainty quantification. 
Moreover, in out-of-distribution tasks, the uncertainty for all the trajectories is high, leading to rejecting all the transitions.

\paragraph{Reconstruction}
UNICORN \cite{unicorn} utilizes a decoder to predict the reward function and next state conditioned on the learned task representations, state, and action. 
Since the reward and transition dynamics are task-related components, the reconstruction objective encourages the context encoder to embed task-related characteristics.

\paragraph{Mutual information}
The CSRO \citep{csro} method is most similar to ours as it implicitly minimizes an upper bound estimate of mutual information between the task representations and the behavior policy. CSRO computes the upper bound based on \citet{club}, by approximating the conditional distribution of the task representations given a behavior policy $p(z_i|s, a)$ as Gaussian. 
In contrast, we show that we can minimize the mutual information by maximizing the entropy of the behavior policy conditioned on the task representations. 
We use generative modeling to approximate this without making strong assumptions about the underlying distribution. 

\section{Method}
In this section, we present our context-based OMRL algorithm, named \textit{Entropy Regularized Task Representation Learning} (\our).
We start by providing a general overview and then we detail our method for learning the task representation. 
In particular, we detail how we reduce the context distribution shift via our novel use of GANs.
Refer to \cref{alg:training} and \cref{fig:overview} in \cref{sec:implementation_detail} for an overview of \our.

\paragraph{Overview} \our has five main components which we wish to learn:
\begin{align}
&\text{Context enc.: } & z_i &=  \mathbb{E}_{(s,a,r,s') \sim c_i} [e_{\theta} (s,a,r,s')] \label{eq:encoder} \\
&\text{Critic: } & q &= Q_{\omega} (s, z_i, a) \label{eq:critic} \\
&\text{Actor: } & a &\sim \pi_{\phi} (s, z_i) \label{eq:actor} \\
&\text{Generator: } & a^{\text{fake}} &= G_{\psi} (s, z_i, \epsilon) \label{eq:generator} \\
&\text{Discriminator: } & p(a) &= D_{\zeta} (a, s, z_i) \label{eq:discriminator}
\end{align}
For a particular task $i$, our encoder $e_{\theta}$ infers the task representations $z_i$ from the context $c_i = \{ (s, a, r, s')^i \}_{j=1}^{H}$, where $H$ is the number of transitions in the context $c_i$.
Note that the $H$ transitions are not necessarily sequential.
In contrast to previous methods, our task representation learning utilizes a GAN -- consisting of a generator $G_{\psi} (s, z_i, \epsilon)$ and discriminator $D_{\zeta} (a, s, z_i)$ -- to reduce context distribution shift, via approximately minimizing the mutual information.
See \cref{sec:task-rep}, for more details.
Once we have inferred the task representations $z_i$, \our follows standard OMRL methodology and conditions the actor $\pi_{\phi}(s,z_i)$ and critic $Q_{\omega}(s, z_i, a)$ on it.

We find that training GAN more often is beneficial since the context encoder is updated at each iteration, leading to changes in task representations $z_i$. 
We also bound the task representations $z_i$ by using Tanh as the activation function of the context encoder.
\subsection{Task Representation Learning} \label{sec:task-rep}
The encoder in \cref{eq:encoder} learns to map the context $c_i$ to task representations $z_i$. 
We train our encoder to minimize
\begin{align} \label{eq:context_objective}
\mathcal{L}(\theta) = \mathcal{L}_{\text{MI}}(\theta) + \lambda \mathcal{L}_{\text{DML}}(\theta),
\end{align}
where $\mathcal{L}_{\text{MI}}(\theta)$ (\cref{eq:mi_loss}) is a term to reduce the amount of context distribution shift, $\mathcal{L}_{\text{DML}}(\theta)$ (\cref{eq:focal}) is a term to preserve distance in the embedding space through distance metric learning and $\lambda$ is a hyper-parameter for balancing these objectives. 
We will now motivate each of these terms and detail how we calculate them in practice.

\subsubsection{Reducing Context Distribution Shift via Mutual Information Minimization \label{sec:mi}}
We seek our context encoder to learn task representations $z_i$ that is independent of the task-specific behavior policy $\pi^i_{\beta}$ which collected the offline data.
The mutual information between two random variables indicates the amount of information obtained from one random variable after observing the other. 
As such, minimizing the mutual information between our task representations $z_i$ and the task-specific behavior policy $\pi_{\beta}^i$ 
would achieve our goal of learning task representations that are independent of the task-specific behavior policy.
However, calculating mutual information 
is not trivial as the task-specific behavior policy $\pi^i_\beta$ is unknown.

\begin{algorithm}[t]
\caption{Meta-training}
\label{alg:training}
\textbf{Input}: Offline datasets $\mathcal{D}= \{ \mathcal{D}_i \} _{i=1}^N$ , context encoder $e_\theta$, generator $G_\psi$, discriminator $D_\zeta$, actor $\pi_\phi$, critic $Q_\omega$
\begin{algorithmic}[1] 
\WHILE{not done}
\STATE Sample context $\{ c_i \} _{i=1}^{H}$ and training data $\{ (s_j, a_j, r_j, s^\prime_j )^i \}_{j=1}^n$ from dataset $\mathcal{D}_i$.
\STATE Obtain task representations $z_i$ according to \cref{eq:encoder}
\FOR{$k=1,2,\ldots,\text{Num}_{\text{Update-GAN}}$}
    \STATE $\epsilon \sim \mathcal{N}(0, \mathbb{I})$
    \STATE Generate actions $a^{\text{fake}}$ according to \cref{eq:generate}.
    \STATE Update discriminator $D_\zeta$  and generator $G_{\psi}$ according to \cref{eq:discrimantor_objective} and \cref{eq:generator_objective}.
\ENDFOR
\STATE $\epsilon \sim \mathcal{N}(0, \mathbb{I})$
\STATE Generate actions $a^{\text{fake}}$ according to \cref{eq:generate}.
\STATE Compute covariance of the  $a^{\text{fake}}$.
\STATE Compute $\mathcal{L}_{\text{MI}}(\theta)$ according to \cref{eq:mi_loss}.
\STATE Compute $\mathcal{L}_{\text{DML}}(\theta)$ according to \cref{eq:focal}.
\STATE Update context encoder $e_\theta$ according to \cref{eq:context_objective}.
\STATE Update the critic $Q_\omega$ according to \cref{eq:critic}.
\STATE Update the actor $\pi_\phi$ according to \cref{eq:actor}.
\ENDWHILE
\STATE \textbf{return} context encoder $e_\theta$ and actor $\pi_\phi$
\end{algorithmic}
\end{algorithm}

We now detail our approach to minimizing the mutual information.
Our approach builds upon the fact that we can rewrite the mutual information in terms of entropies and then utilize a GAN to approximate the objective.
We first note that minimizing the mutual information between each task's representations and its associated behavior policy is equivalent to maximizing the conditional entropy
\begin{equation} \label{eq:entropy}
    \min_\theta \frac{1}{N} \sum_{i=1}^N I(z_i, \pi^i_\beta) = \max_\theta \frac{1}{N} \sum_{i=1}^N H(\pi^i_\beta(a|s)|p(z_i)).
\end{equation}
This is because the mutual information can be written in terms of entropies $I(z_i, \pi^i_\beta) = H(\pi^i_\beta)-H(\pi^i_\beta|p(z_i))$
and as the entropy of the behavior policy $H(\pi^i_\beta)$ is independent of the encoder's parameters $\theta$, the behavior policy's entropy $H(\pi^i_\beta)$ is a constant.  

Whilst we cannot compute $H(\pi^i_\beta(a|s)|p(z_i))$, we can utilize a GAN to estimate the meta-behavior policy $\hat{\pi}_\beta (a|s, z_i)$ by generating actions similar to those in the offline data. 
Importantly, this allows us to approximate the conditional entropy in \cref{eq:entropy} with the entropy of the meta-behavior policy
\begin{equation} \label{eq:entropy_approx}
    H(\pi^i_\beta(a|s)|p(z_i)) \approx H(\hat{\pi}_\beta (a|s, z_i)),
\end{equation}
which is derived in \cref{sec:proof}
The GAN consists of a generator and discriminator, where the generator  samples actions resembling the meta-behavior policy given an observation $s$, task representations $z_i$, and a noise vector $\epsilon$ 
\begin{equation} \label{eq:generate} 
    a^{\text{fake}} = G_\psi (s, z_i, \epsilon).
\end{equation}
The discriminator $p \bigl( a \sim G_\psi (s, z_i, \epsilon) \bigr) = D_\zeta(a, s, z_i)$ then distinguishes between actions in the datasets and the generated actions, where $p \bigl( a  \bigr)$ is the probability of action $a$ being generated. 
To differentiate between the fake (generated) and real (from datasets) actions, the discriminator is trained according to
\begin{equation} \label{eq:discrimantor_objective}
    \mathcal{L}_D(\zeta) = -\log  D_\zeta(a^{\text{real}}, s, z_i) -\log{(1- D_\zeta(a^{\text{fake}}, s, z_i))}.
\end{equation}
On the other hand, the generator is trained to fool the discriminator by generating fake actions similar to the actions in the dataset:
\begin{equation} \label{eq:generator_objective}
    \mathcal{L}_G(\psi) = \log  D_\zeta(G_\psi (s, z_i, \epsilon), s, z_i).
\end{equation}
As training progresses, the distribution of the generated samples should better reflect the underlying meta-behavior policy. 
We generate samples from the meta-behavior policy to decrease context shift by maximizing the entropy based on the task representation. 
We assume a multivariate Gaussian distribution for samples and compute the covariance matrix of the generated samples
$\Sigma_{j,k} = \mathbb{E}[(a^{\text{fake}}_j-\mathbb{E}[a^{\text{fake}}_j])(a^{\text{fake}}_k-\mathbb{E}[a^{\text{fake}}_k])]$ and determine the entropy accordingly:
\begin{equation} \label{eq:entropy_samples}
    H(\hat{\pi}_\beta (a|s, z_i)) \approx \frac{1}{2} \log \det {\Sigma} + \frac{k}{2} \log{(2\pi e)},
\end{equation}
where $k$ is the dimensionality of the action space. 
As stated, maximizing the entropy of samples is equivalent to minimizing the mutual information between task representations and behavior policy, leading to a decrease in context shift. 
This gives our context encoder's loss term
\begin{equation} \label{eq:mi_loss}
    \mathcal{L}_{\text{MI}}(\theta) = -\frac{1}{2} \log \det \Sigma. 
\end{equation}
To avoid numerical instability, we add a small constant $10^{-5}$ to diagonal elements of the covariance matrix.

\subsubsection{Distance Metric Learning}

Ideally, the context encoder should preserve distances when embedding, that is, it should embed contexts from the same task close together and contexts from different tasks further apart. 
This enables the agent to distinguish between different tasks and adapt accordingly.  
We use distance metric similar to FOCAL \citep{focal} to achieve distinct task representations:
\begin{multline} \label{eq:focal}
    \mathcal{L}_{\text{DML}}(\theta) = \mathbf{1} \{ i=j \} || z_i - z_j||^2_2 
    \\ 
    + \mathbf{1} \{ i \ne j \} \frac{\beta}{|| z_i - z_j||^2_2 + \epsilon_0},
\end{multline}
where $\epsilon_0$ is a hyper-parameter added to avoid zero division.
It has been shown that this objective approximately maximizes the mutual information between the task representations and the corresponding task \citep{csro, unicorn}. 
\cref{fig:ablation} in \cref{sec:ablation} illustrates the necessity of this objective.
\begin{table*}[t]
\centering
\begin{tabular}{c|c|ccccc} 
\hline\hline
\multicolumn{1}{l}{\sc Environment} & \multicolumn{1}{l}{\sc Context}  & {\sc offlinepearl} & {\sc focal} & \sc{csro} & {\sc unicorn} & \our (Ours) \\ 
\hline
 Cheetah-Vel     & \multirow{8}{*}{Offline} & $ 66.19 \pm 14.41 $ & $ 69.33 \pm 0.99 $  & $ 66.48 \pm 2.66 $ & $ 66.44 \pm 1.55 $ & $ \textbf{72.61} \pm 1.39 $  \\
 Ant-Goal        &                          & $ \textbf{82.16} \pm 4.71 $  & $ 61.37 \pm 5.56 $  & $ 52.05 \pm 3.68 $ & $ 61.60 \pm 3.37 $ & $ 79.14 \pm 4.15 $  \\
 Ant-Dir         &                          & $ -18.87 \pm 3.15 $ & $ 47.75 \pm 3.31 $  & $ 51.51 \pm 3.99 $ & $ 50.23 \pm 2.34 $ & $ \textbf{54.86} \pm 2.82 $  \\
 Humanoid-Dir    &                          & $ \textbf{61.30} \pm 6.94 $  & $ 49.13 \pm 2.65 $  & $ 48.94 \pm 2.82 $ & $ 44.82 \pm 2.28 $ & $ 59.93 \pm 2.79 $  \\
 Hopper-Mass     &                          & $ 84.13 \pm 15.47 $ & $ 80.56 \pm 14.54 $ & $ 80.23 \pm 7.53 $ & $ 83.88 \pm 8.10 $ & $ \underline{\textbf{99.72}} \pm 0.86 $  \\
 Hopper-Friction &                          & $ 51.61 \pm 10.02 $ & $ 57.77 \pm 7.18 $  & $ 63.12 \pm 8.93 $ & $ 56.77 \pm 7.55 $ & $ \textbf{67.44} \pm 6.71 $  \\
 Walker-Mass     &                          & $ 36.97 \pm 2.60 $  & $ 35.36 \pm 5.27 $  & $ 43.21 \pm 5.31 $ & $ 39.61 \pm 8.82 $ & $ \underline{\textbf{61.73}} \pm 2.15 $  \\
 Walker-Friction &                          & $ 60.15 \pm 5.35 $  & $ 39.54 \pm 6.48 $  & $ 49.37 \pm 5.14 $ & $ 46.18 \pm 6.84 $ & $ \textbf{62.78} \pm 10.89 $ \\
\hline
 Cheetah-Vel     & \multirow{8}{*}{Online}  & $ 56.24 \pm 10.60 $ & $ 54.94 \pm 3.50 $  & $ 50.57 \pm 11.06 $ & $ 51.61 \pm 3.17 $  & $ \textbf{61.85} \pm 2.18 $ \\
 Ant-Goal        &                          & $ 20.07 \pm 4.98 $  & $ 29.23 \pm 5.59 $  & $ 26.45 \pm 3.57 $  & $ 26.31 \pm 4.91 $  & $ \underline{\textbf{49.15}} \pm 4.30 $ \\
 Ant-Dir         &                          & $ -19.91 \pm 3.97 $ & $ 4.01 \pm 3.60 $   & $ 8.52 \pm 6.63 $   & $ 4.50 \pm 7.95 $   & $ \underline{\textbf{27.63}} \pm 5.46 $ \\
 Humanoid-Dir    &                          & $ \textbf{59.43} \pm 7.00 $  & $ 44.37 \pm 3.72 $  & $ 46.67 \pm 5.82 $  & $ 36.65 \pm 1.83 $  & $ 53.39 \pm 3.33 $ \\
 Hopper-Mass     &                          & $ 77.36 \pm 12.14 $ & $ 81.61 \pm 9.33 $  & $ 84.80 \pm 10.06 $ & $ 83.36 \pm 5.77 $  & $ \underline{\textbf{99.59}} \pm 1.39 $ \\
 Hopper-Friction &                          & $ 48.04 \pm 10.95 $ & $ 56.93 \pm 12.62 $ & $ 60.95 \pm 9.26 $  & $ 52.77 \pm 16.28 $ & $ \textbf{66.35} \pm 8.73 $ \\
 Walker-Mass     &                          & $ 34.48 \pm 6.46 $  & $ 39.10 \pm 4.82 $  & $ 39.70 \pm 5.90 $  & $ 38.29 \pm 7.51 $  & $ \underline{\textbf{65.15}} \pm 4.48 $ \\
 Walker-Friction &                          & $ 41.73 \pm 9.48 $  & $ 45.51 \pm 5.78 $  & $ 49.47 \pm 9.21 $  & $ 47.18 \pm 7.21 $  & $ \textbf{58.80} \pm 9.94 $ \\
\hline\hline
\end{tabular}
\caption{\textbf{Improved out-of-distribution generalization} The average normalized return for out-of-distribution test tasks after 100k training steps, averaged over 5 random seeds, $\pm$ represents standard deviation. 
\textbf{Bold} indicate highest mean value and \textbf{\underline{underline}} indicate statistical significance according to t-test with p-value $< 0.05$.}
\label{tab:ood_eval}
\end{table*}

\begin{table*}[t]
\centering
\setlength{\tabcolsep}{1mm}
\begin{tabular}{c|c|ccccc} 

\hline\hline
\multicolumn{1}{l}{\sc Environment} & \multicolumn{1}{l}{\sc Model}  & {\sc offlinepearl} & {\sc focal} & \sc{csro} & {\sc unicorn} & \our (Ours) \\ 
\hline
 Cheetah-Vel     & \multirow{8}{*}{\begin{tabular}[c]{@{}c@{}}Linear\\ 
                             Regression\end{tabular}} & $ 0.3451 \pm 0.0040 $ & $ 0.1720 \pm 0.0014 $ & $ \underline{\textbf{0.1586}} \pm 0.0009 $ & $ 0.1804 \pm 0.0013 $ & $ 0.1791 \pm 0.0017 $ \\
 Ant-Goal        &                                    & $ 0.6477 \pm 0.0029 $ & $ 0.5229 \pm 0.0015 $ & $ 0.6235 \pm 0.0030 $ & $ 0.4996 \pm 0.0035 $ & $ \underline{\textbf{0.4617}} \pm 0.0048 $ \\
 Ant-Dir         &                                    & $ 1.6278 \pm 0.5446 $ & $ 0.6144 \pm 0.0066 $ & $ 0.4895 \pm 0.0123 $ & $ 0.5313 \pm 0.0129 $ & $ \underline{\textbf{0.4546}} \pm 0.0171 $ \\
 Humanoid-Dir    &                                    & $ 1.3596 \pm 0.0000 $ & $ 0.8661 \pm 0.0140 $ & $ 0.9876 \pm 0.0081 $ & $ 0.8760 \pm 0.0047 $ & $ \textbf{0.8537} \pm 0.0010 $ \\
 Hopper-Mass     &                                    & $ 0.3488 \pm 0.0001 $ & $ 0.3218 \pm 0.0012 $ & $ 0.2644 \pm 0.0023 $ & $ 0.2383 \pm 0.0030 $ & $ \underline{\textbf{0.2265}} \pm 0.0026 $ \\
 Hopper-Friction &                                    & $ 0.3661 \pm 0.0001 $ & $ \underline{\textbf{0.2443}} \pm 0.0028 $ & $ 0.2586 \pm 0.0035 $ & $ 0.2518 \pm 0.0015 $ & $ 0.2551 \pm 0.0015 $ \\
 Walker-Mass     &                                    & $ 0.3533 \pm 0.0012 $ & $ 0.3035 \pm 0.0028 $ & $ 0.2895 \pm 0.0020 $ & $ \underline{\textbf{0.2712}} \pm 0.0006 $ & $ 0.2794 \pm 0.0026 $ \\
 Walker-Friction &                                    & $ 0.3696 \pm 0.0006 $ & $ 0.3715 \pm 0.0011 $ & $ 0.3640 \pm 0.0011 $ & $ 0.3679 \pm 0.0014 $ & $ \underline{\textbf{0.3526}} \pm 0.0015 $ \\
\hline
 Cheetah-Vel     & \multirow{8}{*}{SVR}               & $ 0.3643 \pm 0.0018 $ & $ 0.1696 \pm 0.0011 $ & $ \underline{\textbf{0.1656}} \pm 0.0013 $ & $ 0.1657 \pm 0.0013 $ & $ 0.1689 \pm 0.0009 $ \\
 Ant-Goal        &                                    & $ 0.6528 \pm 0.0017 $ & $ 0.4998 \pm 0.0021 $ & $ 0.5171 \pm 0.0043 $ & $ 0.4850 \pm 0.0025 $ & $ \underline{\textbf{0.4182}} \pm 0.0022 $ \\
 Ant-Dir         &                                    & $ 1.3561 \pm 0.0006 $ & $ 0.4832 \pm 0.0067 $ & $ 0.4753 \pm 0.0082 $ & $ 0.4684 \pm 0.0057 $ & $ \underline{\textbf{0.4421}} \pm 0.0105 $ \\
 Humanoid-Dir    &                                    & $ 1.3596 \pm 0.0000 $ & $ 0.8242 \pm 0.0058 $ & $ 0.8878 \pm 0.0039 $ & $ \textbf{0.8163} \pm 0.0060 $ & $ 0.8186 \pm 0.0051 $ \\
 Hopper-Mass     &                                    & $ 0.3453 \pm 0.0010 $ & $ 0.3220 \pm 0.0011 $ & $ 0.2486 \pm 0.0016 $ & $ 0.2362 \pm 0.0007 $ & $ \underline{\textbf{0.2146}} \pm 0.0011 $ \\
 Hopper-Friction &                                    & $ 0.3649 \pm 0.0011 $ & $ 0.2316 \pm 0.0021 $ & $ 0.2443 \pm 0.0015 $ & $ \underline{\textbf{0.2285}} \pm 0.0012 $ & $ 0.2328 \pm 0.0017 $ \\
 Walker-Mass     &                                    & $ 0.3499 \pm 0.0009 $ & $ 0.2812 \pm 0.0006 $ & $ 0.2802 \pm 0.0010 $ & $ 0.2656 \pm 0.0011 $ & $ \underline{\textbf{0.2515}} \pm 0.0007 $ \\
 Walker-Friction &                                    & $ 0.3713 \pm 0.0004 $ & $ 0.3749 \pm 0.0007 $ & $ 0.3618 \pm 0.0004 $ & $ 0.3545 \pm 0.0009 $ & $ \underline{\textbf{0.3301}} \pm 0.0006 $ \\
\hline\hline
\end{tabular}
\caption{\textbf{Better task representation} The RMSE for predicting the true labels based on the task representations for test samples, 
\textbf{Bold} indicate lowest mean value and \textbf{\underline{underline}} indicate statistical significance according to t-test with p-value $< 0.05$.}
\label{tab:predictions}
\end{table*}

\subsection{Meta-RL with Task Representations}
Given the task representation learning strategy in \cref{sec:task-rep}, \our follows standard OMRL methodologies and conditions the agent on the task representations $z_i$ so that it can adapt accordingly. 
We utilize the actor-critic framework to train the agent using temporal difference (TD) learning from offline datasets. 
Distribution shift \citep{levine_tutorial} is a known problem in offline RL: $Q$ values are overestimated because of a difference between the behavior policy and the policy being learned.
We adopt BRAC \cite{brac} as the base offline RL algorithm which regularizes policy to alleviate issues arising from distribution shift when learning the actor and critic.
The critic is trained to minimize
\begin{align} \label{eq:critic}
    \mathcal{L}_{Q}(\omega; s,a,r,s^\prime, z_i) &= \left( Q_\omega(s,a,z_i)- y \right)^2 \\
    y&=r-\gamma Q_{\bar{\omega}}(s^\prime, a^\prime, z_i),
\end{align}
where the task representations $z_i$ is inferred by the encoder in \cref{eq:encoder}, the next action is given by $a^\prime \sim \pi_\phi(.|s^\prime, z_i)$ and the target $y$ uses the exponential moving average (EMA) of the critic's weights, i.e. $\bar{\omega} = \tau \omega + (1-\tau) \bar{\omega}$.
The actor is then trained to minimize
\begin{align} \label{eq:actor}
    \mathcal{L}_{\pi}(\phi; s, z_i) &= Q_\omega(s, \tilde{a}, z_i)-\alpha D_{\text{KL}} \left( \pi_\phi(.|s,z_i) || \pi^i_\beta (.|s) \right),
\end{align}
where the action is a sample from the policy ${\tilde{a} \sim \pi_{\phi}(a \mid s, z_i)}$, $D_{\text{KL}}$ is the estimate of the KL divergence between the learned policy $\pi_\phi$ and the behavior policy $\pi_\beta$, and $\alpha$ is a hyper-parameter for the amount of regularization. We used the dual form of KL divergence according to BRAC \cite{brac}.

\section{Experiments}
In this section, we evaluate \our in a set of multi-task MuJoCo environments \cite{mujoco}. Our experiments seek to answer the following questions:
\begin{enumerate}
    \item Does \our's method for learning task representations improve adaptation performance and generalization?
    \item Does our method for decoupling task representation learning from datasets capture the underlying task?
    \item When does reducing the context shift improve the performance of OMRL agents?
\end{enumerate}
\paragraph{Environments}
We now detail the environments and the dynamics parameters which we considered to be in-distribution and out-of-distribution.
Note that we train on in-distribution parameters and evaluated on both in- and out-of-distribution parameters.
\begin{itemize}
    \item \textbf{Cheetah-Vel}: a cheetah robot moves forward at a target velocity. The velocity for in-distribution tasks is between $[1, 2]$ and for out-of-distribution tasks is between $[0.5, 1]$ and $[2, 2.5]$.
    \item \textbf{Ant-Goal}: an ant robot moves to reach a goal state. The goals are in a semi-circle where the radius for in-distribution tasks is either $0.8$ or $1.2$ and for out-of-distribution tasks is $1.6$, which is further. 
    \item \textbf{Ant-Dir} and \textbf{Humanoid-Dir}: an ant/humanoid robot moves in a certain direction.  The directions for in-distribution tasks is between $[-\frac{\pi}{2}, \frac{\pi}{2}]$ and for out-of-distribution tasks is between $[-\frac{3\pi}{4}, -\frac{\pi}{2}]$ and $[\frac{\pi}{2}, \frac{3\pi}{4}]$.
    \item \textbf{Hopper-Mass}, \textbf{Walker-mass}, \textbf{Hopper-Friction}, and \textbf{Walker-Friction}: a hopper (one-legged robot) or walker (bi-legged) robot must move as fast as it can while either the mass of the links or the friction varies in each task. The original mass and friction are scaled by a factor.  For in-distribution tasks the factor is between $[1.5^{-1}, 1.5^{1}]$ and for out-of-distribution tasks the factor is between $[1.5^{-1.5}, 1.5^{-1}]$ and $[1.5^{1}, 1.5^{1.5}]$. 
\end{itemize}
In the first 4 environments, the reward function is different for each task and the transition dynamics is different in the other 4 environments. 
To test the generalization of OMRL methods, we use different task distributions for training and testing and we divide the testing tasks into \textit{in-distribution} and \textit{out-of-distribution} tasks. 
In each environment, we consider 20 tasks for training, 10 tasks for in-distribution testing, and 10 tasks for out-of-distribution testing. 
See \cref{tab:envs} in \cref{sec:env} for more details about the environments. 

\begin{table*}[t]
\centering
\begin{tabular}{c|c|ccccc} 

\hline\hline
\multicolumn{1}{l}{\sc Environment} & \multicolumn{1}{l}{\sc Context}  & {\sc offlinepearl} & {\sc focal} & \sc{csro} & {\sc unicorn} & \our (Ours) \\ 
\hline
 Cheetah-Vel     & \multirow{8}{*}{Offline} & $ 83.30 \pm 20.31 $ & $ 92.84 \pm 1.40 $  & $ 94.01 \pm 0.74 $  & $ 93.53 \pm 0.80 $  & $ \textbf{96.36} \pm 0.19 $  \\
 Ant-Goal        &                          & $ 98.97 \pm 3.69 $  & $ 93.90 \pm 2.83 $  & $ 94.62 \pm 3.67 $  & $ 95.53 \pm 2.71 $  & $ \underline{\textbf{107.66}} \pm 2.50 $ \\
 Ant-Dir         &                          & $ 28.66 \pm 2.99 $  & $ 60.75 \pm 1.74 $  & $ 65.20 \pm 4.66 $  & $ 62.29 \pm 1.95 $  & $ \textbf{69.08} \pm 4.64 $  \\
 Humanoid-Dir    &                          & $ 66.26 \pm 4.19 $  & $ 62.14 \pm 1.95 $  & $ 56.33 \pm 1.36 $  & $ 58.02 \pm 2.24 $  & $ \textbf{70.53} \pm 5.54 $  \\
 Hopper-Mass     &                          & $ 89.45 \pm 9.44 $  & $ 79.42 \pm 10.97 $ & $ 83.74 \pm 12.11 $ & $ 85.84 \pm 9.63 $  & $ \underline{\textbf{98.73}} \pm 0.32 $  \\
 Hopper-Friction &                          & $ 79.44 \pm 11.90 $ & $ 80.85 \pm 14.87 $ & $ 82.60 \pm 9.98 $  & $ 80.55 \pm 13.39 $ & $ \textbf{86.65} \pm 15.66 $ \\
 Walker-Mass     &                          & $ 52.80 \pm 2.30 $  & $ 54.09 \pm 4.76 $  & $ 55.49 \pm 1.28 $  & $ 56.68 \pm 2.47 $  & $ \underline{\textbf{77.52}} \pm 1.39 $  \\
 Walker-Friction &                          & $ 56.48 \pm 4.78 $  & $ 47.76 \pm 4.52 $  & $ 48.89 \pm 6.18 $  & $ 51.05 \pm 2.39 $  & $ \textbf{64.22} \pm 6.96 $  \\
\hline
 Cheetah-Vel     & \multirow{8}{*}{Online}  & $ 80.04 \pm 19.24 $ & $ 85.98 \pm 1.79 $  & $ 89.35 \pm 1.79 $  & $ 88.16 \pm 1.67 $  & $ \textbf{92.47} \pm 0.36 $  \\
 Ant-Goal        &                          & $ 49.07 \pm 3.22 $  & $ 77.15 \pm 3.17 $  & $ 80.86 \pm 4.39 $  & $ 79.04 \pm 2.11 $  & $ \underline{\textbf{93.07}} \pm 1.95 $  \\
 Ant-Dir         &                          & $ 29.09 \pm 3.82 $  & $ 42.43 \pm 5.18 $  & $ \textbf{54.64} \pm 3.89 $  & $ 41.25 \pm 5.11 $  & $ 53.56 \pm 11.56 $ \\
 Humanoid-Dir    &                          & $ \textbf{66.95} \pm 5.71 $  & $ 54.40 \pm 1.91 $  & $ 52.13 \pm 3.08 $  & $ 44.90 \pm 2.13 $  & $ 61.36 \pm 2.42 $  \\
 Hopper-Mass     &                          & $ 88.18 \pm 8.82 $  & $ 86.47 \pm 8.19 $  & $ 88.85 \pm 3.65 $  & $ 84.38 \pm 12.25 $ & $ \underline{\textbf{99.66}} \pm 1.11 $  \\
 Hopper-Friction &                          & $ 65.92 \pm 13.60 $ & $ 74.31 \pm 30.60 $ & $ 81.91 \pm 14.96 $ & $ 80.49 \pm 14.00 $ & $ \textbf{92.93} \pm 6.38 $  \\
 Walker-Mass     &                          & $ 36.84 \pm 8.80 $  & $ 39.23 \pm 4.42 $  & $ 46.31 \pm 8.94 $  & $ 37.15 \pm 10.26 $ & $ \underline{\textbf{70.73}} \pm 2.32 $  \\
 Walker-Friction &                          & $ 36.52 \pm 9.27 $  & $ 39.24 \pm 5.51 $  & $ 46.43 \pm 8.97 $  & $ 38.42 \pm 6.72 $  & $ \textbf{55.64} \pm 11.06 $ \\
\hline\hline
\end{tabular}
\caption{\textbf{Better in-distribution performance} The average normalized return for in-distribution tasks after 100k training steps, averaged over 5 random seeds, $\pm$ represents standard deviation. 
\textbf{Bold} indicate highest mean value and \textbf{\underline{underline}} indicate statistical significance according to t-test with p-value $< 0.05$.}
\label{tab:id_eval}
\end{table*}

\paragraph{Baselines}
To evaluate the performance of our method, we compare it to the following baselines: 
\begin{itemize}
    \item \textbf{FOCAL} \citep{focal} which utilizes distance metric objective to train the context encoder.
    \item \textbf{CSRO} which minimizes an upper bound of mutual information using CLUB \citep{club}.  
    \item \textbf{UNICORN} \citep{unicorn} which utilizes a decoder to predict the next state, reward, and distance metric.
    \item \textbf{OfflinePEARL} which is an offline extension of PEARL \citep{pearl} where the context encoder is trained based on the objective of the value function end-to-end.
\end{itemize}
We use the same network architecture, offline RL algorithm, and common hyper-parameters to compare fairly and focus on task representation learning.

\paragraph{Offline data}
We used soft actor-critic (SAC, \citealp{sac}) to generate the datasets and trained each agent to an expert level. 
The dataset consists of trajectories collected from rolling out the corresponding SAC agent at different training stages; each dataset contains 180k transitions. 

Please see \cref{sec:implementation_detail} for further details of our experiments.
We also provide training curves in \cref{sec:plots} and we report results for unnormalized returns in \cref{sec:unnormalized} because this enables a fair comparison to prior implementations.  

\subsection{\our's Adaptation Performance}
We evaluate the performance of our method alongside baselines on two scenarios. In the first (unrealistic) case, the context is given from the offline dataset collected from the behavior policy. In the second (realistic) case, agents gather the context by interacting with the environment conditioned on a prior task representations $z_0$. 
Refer to \cref{sec:nonprior} for non-prior context collection, introduced by \cite{csro}.

\paragraph{Does \our's tasks representation learning improve performance in out-of-distribution tasks?}
\cref{tab:ood_eval} summarizes the results for out-of-distribution test tasks, where the distribution of tasks is different to training, allowing us to compare generalization.
We normalized the return for each task such that a random agent yields a return of 0 and an expert SAC agent yields a return of 100. 
When the context is given, OfflinePEARL, which is trained end-to-end, has a competitive performance in most environments, even surpassing other methods in several tasks. 
However, when the context is collected by the agent, there is a sharp drop in performance for many environments due to the distribution shift. 
Our method (\our), CSRO, and UNICRON address this issue by reducing the correlation between the task representations and the policy that collected the context. 
However, our method outperformed the others in 7 out of 8 environments and there is a significant difference in 4 of them. 
This indicates that our approach to reducing the distribution shift is more effective. 

\paragraph{Does \our's tasks representation learning improve performance in in-distribution tasks?}
\cref{tab:id_eval} summarizes the results for in-distribution tasks. 
Our method, \our, outperforms baselines in almost all experiments.
For the case of online context collection, OfflinePEARL in Humanoid-Dir and CSRO in Ant-Dir perform slightly better than our method.  
In 3 environments, our method's performance significantly surpasses the others according to the t-test for both offline and online contexts. 

\subsection{Evaluating \our's Task Representation Learning}
To evaluate the quality of the learned task representations in different methods, we trained simple regression models to predict the true goal label. 
For example, in the Ant-Dir environment, the goal label is the direction that the ant should move in the task. 
For each in-distribution and out-of-distribution test task, we sample 1000 transitions and embed them in task representations with the trained context encoder. 
We then use 80\% of the samples for training and 20\% for testing. 
Finally, we utilize linear regression and support vector regression (SVR, \citealp{svr}) with a radial basis function kernel to compare the quality of learned task representations via root mean squared error (RMSE). 
\cref{tab:predictions} compares the quality of different task representations based on RMSE of goal predictions for test samples. 

\subsection{When does mutual information objective improve the performance?}
\begin{figure}[h]
    \centering
    \includegraphics[width=\linewidth]{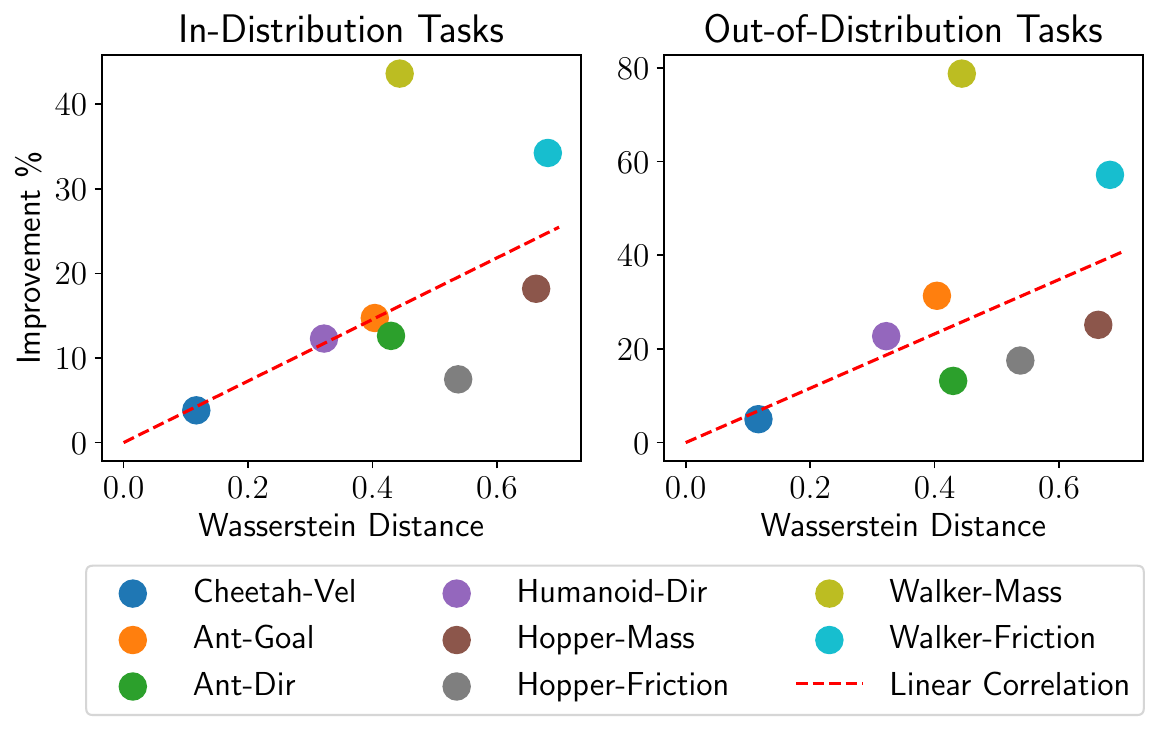}
    \caption{Correlation between the distance between expert policies and performance improvement by utilizing mutual information objective. For ID tasks, the Pearson coefficient is $\rho=0.75$ with $p=0.058$, and for OOD tasks $\rho=0.69$ with $p=0.086$.  
    }
    \label{fig:mi_distance}
\end{figure}

The behavior policy affect the context during training and the context encoder may embed the characteristics of the behavior policy in the task representations. 
We addressed this issue by minimizing the mutual information between the task representations and the behavior policy. 
An interesting question is: when does this objective improve performance? 
Our intuition is that this objective improves the performance for environments where the behavior policies are different for each task. 
Therefore, the task representations should not embed the characteristics of the behavior policy. 
To evaluate our intuition, we compute the Wasserstein distance between the expert policies (SAC agents) used for data collection. 
These agents assumed a Gaussian distribution over the action given an observation.
The  Wasserstein distance between Gaussian distributions is defined as:
\begin{multline} \label{eq:def_wasserstein}
    W_2 (\mathcal{N}(\mu_1, \Sigma_1); \mathcal{N}(\mu_2, \Sigma_2)) =  
    \\ 
    \sqrt{||\mu_1-\mu_2||^2_2 + \mathrm{Tr} \bigl( \Sigma_1 + \Sigma_2 - 2(\Sigma_1^{1/2}\Sigma_2 \Sigma_1^{1/2})^\frac{1}{2} \bigr)}.
\end{multline}
We consider the average Wasserstein distance between the expert policies as an estimate of their difference 
\begin{equation}
    d = \frac{\sum_{i=1}^n \sum_{j=1}^n W_2(\pi_\beta^i ; \pi_\beta^j)}{kn^2},
\end{equation}
where $n$ is the number of tasks and $k$ is dimensionality of action space. 
We divide the distance by the dimensionality of the action space since different environments have different action spaces and the Wasserstein distance depends on that. 
\cref{fig:mi_distance} illustrates the relationship between this distance and the improvement in performance when the mutual information objective (\cref{eq:mi_loss}) is considered for different environments. 
There is a correlation between the average Wasserstein distance and the improvement in performance for both in-distribution tasks and out-of-distribution tasks.

\subsection{Latent Space Visualization}
\begin{figure}[h]
    \centering
    \includegraphics[width=\linewidth]{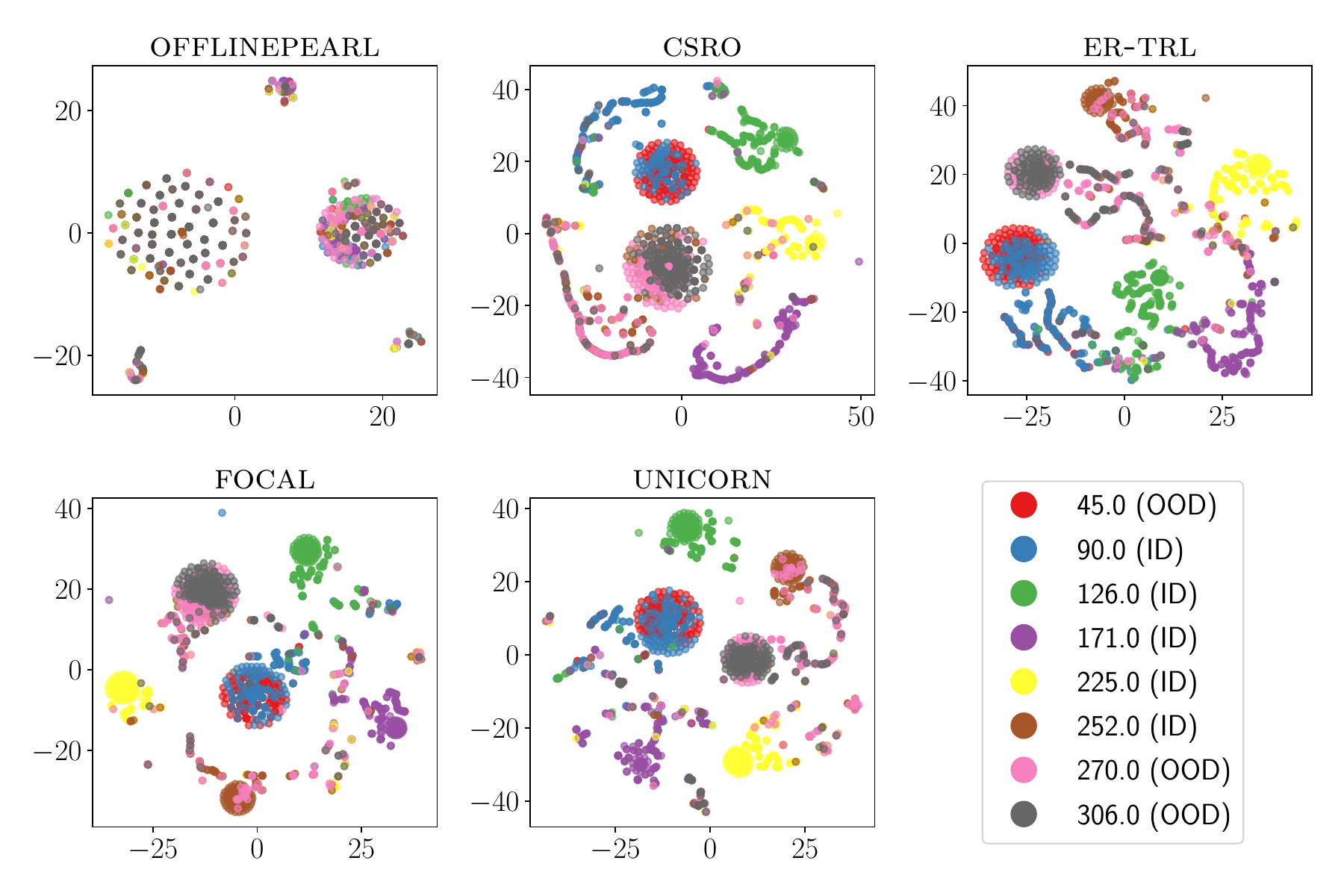}
    \caption{\textbf{Distinguishable task representation learning} Visualizing task representations with t-SNE projection for Ant-Dir environment. 
    The labels illustrate the direction in which the ant robot should move in degrees.}
    \label{fig:latent_main_antdir}
\end{figure}

The main focus of context-based OMRL is on task representation learning. 
We visualize the task representations for 5 in-distribution tasks and 3 out-of-distribution test tasks. 
We use 256 samples for each task, compute the task representations for the same samples with different methods, and project the task representations into a two-dimensional space using t-SNE \citep{tsne}. 
\cref{fig:latent_main_antdir} illustrates the projection of the task representations for the Ant-Dir environment. 
OfflinePEARL fails to learn good representations and task representations for different tasks cannot be visually distinguished. 
\our learns distinguishable task representations while preserving the distance in the embedding space between different tasks. 
We provide further visualizations in \cref{sec:tsne}, \cref{fig:tsne_reward} and \cref{fig:tsne_dynamics}.

\section{Conclusion}
This paper presents a novel approach to deal with context distribution shift in offline meta-RL, where the context encoder overfits to the dataset by embedding behavior-related characteristics. 
We propose to indirectly minimize the mutual information between the task representations and the behavior policy, we show that this is equivalent to maximizing the entropy of a meta-behavior policy. 
We utilized generative modeling, specifically a GAN, to generate actions resembling the meta-behavior policy and subsequently trained the context encoder to maximize the entropy of the generated actions. 
Experiments show that including this objective when training the context encoder captures the underlying tasks more accurately by learning task representations focused on embedding task-related characteristics. 
Subsequently, our entropy-regularized task representation learning outperformed previous context-based OMRL methods in both in-distribution and out-of-distribution tasks, improving online adaptation and generalization. 
\section*{Acknowledgements}
We acknowledge CSC – IT Center for Science, Finland, for awarding this project
access to the LUMI supercomputer, owned by the EuroHPC Joint Undertaking, hosted by CSC
(Finland) and the LUMI consortium through CSC. We acknowledge the computational resources
provided by the Aalto Science-IT project. 
J.~Pajarinen was partly supported by Research Council of Finland (345521). M.~Nakhaei was supported by Business Finland (BIOND4.0 - Data Driven Control for Bioprocesses).
A.~Scannell was supported by the Research Council of Finland from the Flagship program: Finnish Center for Artificial Intelligence (FCAI). 
\bibliography{./aaai25.bib}
\appendix
\section{Proof for Approximating Conditional Entropies \label{sec:proof}}
In \cref{sec:mi}, we approximate the conditional entropies by the entropy of a meta-behavior policy: 
\begin{equation} \label{eq:entropy_approx}
    H(\pi_\beta^i (a|s)| p(z_i)) \approx H(\hat{\pi}_\beta(a|s,z_i)).
\end{equation}
Based on the definition of conditional entropies, we can rewrite \cref{eq:entropy_approx} in the form joint entropies and the entropy of the conditioned random variable:
\begin{equation} \label{eq:convert_cond_entropies}
    H(\pi_\beta^i (a|s)| p(z_i)) = H(\pi_\beta^i (a|s), p(z_i)) - H(p(z_i)).
\end{equation}
Considering the fact that $p(\pi_\beta^i(a|s), p(z_i)) = \hat{\pi}_\beta(a|s,z_i)p(z_i)$ and by expanding the first term (joint entropy) in \cref{eq:convert_cond_entropies}, we can simplify the joint entropies according to:
\begin{multline} \label{eq:derive}
    H(\pi_\beta^i (a|s), p(z_i)) = \\ 
    -\int_a \int_{z_i} \hat{\pi}_\beta(a|s,z_i)p(z_i)  \log {\hat{\pi}_\beta(a|s,z_i)p(z_i)}d_{z_i}  d_a = \\ 
    -\int_a \int_{z_i} \hat{\pi}_\beta(a|s,z_i)p(z_i)  \log {\hat{\pi}_\beta(a|s,z_i)}d_{z_i} \\ 
    -\int_a \int_{z_i} \hat{\pi}_\beta(a|s,z_i)p(z_i)  \log {p(z_i)}d_{z_i} =\\ E_{p(z_i)}[H(\hat{\pi}_\beta(a|s,z_i))] + E_{\hat{\pi}_\beta(a|s,z_i)}[H(p(z_i))]
\end{multline}
Adding \cref{eq:derive} and the second term in \cref{eq:convert_cond_entropies} and simplifying $E_{\hat{\pi}_\beta(a|s,z_i)}[H(p(z_i))] = H(p(z_i))$ result in:
\begin{equation} \label{eq:approx_final}
    H(\pi_\beta^i (a|s)| p(z_i)) = \\E_{p(z_i)}[H(\hat{\pi}_\beta(a|s,z_i))]
\end{equation}
If context encoder is deterministic, as in \our, the approximation in \cref{eq:entropy_approx} becomes equality.

\section{Extended Related Work}
\our is based on offline RL, meta RL, and representation learning. 
We first provide an overview of offline RL algorithms, then we discuss different approaches to meta-RL and the extension to the offline setting. 
Finally, we describe prior research on representation learning in RL.  
\subsection{Offline Reinforcement Learning}
Reinforcement learning enables artificial agents to learn from trial and error and by interacting with the environment. 
Online data collection can be expensive and potential dangerous in real-world applications. 
Offline RL is an emerging paradigm aimed at sidestepping this issue by enabling agents to learn from previously collected static datasets. 
However, datasets usually do not cover the whole state-action space and the learned agents can select actions not included in the datasets. 
When combined with temporal difference (TD) learning, out-of-distribution (OOD) actions can lead to overestimation of value function, limiting the performance and generalization of the agents. 

One approach is to regularize the policy (actor) to be close the the policy that collected the datasets, referred as behavior policy \cite{bcq, td3bc, td3bc_imporved, td7, entropy_diffusion, efficient_diffusion}. 
Another class of methods penalize the value function for OOD actions, learning pessimistic value functions \cite{cql, mcq, combo, edac, cbop, csve}. 
In-sample learning side-steps overestimation of value function by only selecting actions in the dataset for bootstrapping and performing weighted imitation learning \cite{iql, sql, idac_diff, insample_softmax}.
Another class of methods utilize generative modeling \cite{dt, odt, diffuser, dd, mamba_decision}. 
These methods do not rely on bootstrapping and value estimation, overcoming overestimation of the value function.

The performance of the offline agents is limited to the quality of the datasets \cite{dataset}, therefore, fine-tuning the performance of pre-trained agents by allowing to interact with the environment is gaining interest. 
\citet{balanced_replay} utilizes a balanced replay to sample near on-policy transition during fine-tuning stage. 
Adaptive BC \cite{adaptive_bc} adjusts the behavior cloning regularization term in TD3BC \cite{td3bc} inspired by proportional-derivative (PD) controllers. 
PEX \cite{pex} fixes the offline policy and train an additional one while sharing the value function, policies are selected based on their Q values on the fine-tuning phase. 
\citet{q_ensemble_adaptation, robust_ensemble_offline_to_online, perspective_offline_to_online} utilize an ensemble of value functions while considering smoothness to prevent the policy to exploit sub-optimal actions with sharp value functions. 
\citet{relce} proposed a residual policy while encoding prior transitions to enable fast adaptation to different dynamic perturbations in the fine-tuning stage. 

\subsection{Meta-Reinforcement Learning}
Reinforcement learning enables agents to learn a specialized policy for a certain task. 
Given a new similar task, agents need to learn from scratch. 
Meta-RL consider a distribution of training tasks and enables fast adaptation within few trial. 
Gradient-based approaches \cite{maml, meta_model_base} learns the parameters of the networks such that within few gradient steps, the agent can adapt to new tasks by exploiting the common structure in the training tasks. 
Context-based approaches embed the previous transition, referred as context, to task representations to infer the corresponding task. 
The RL agent is then conditioned on task representations to adapt
accordingly. 
PEARL \cite{pearl} learns a temporal-invariant probabilistic context encoder based on the objective of the value function. 
Varibad \cite{varibad} utilize an RNN variational context encoder to infer the task while context encoder is trained separately from the RL agent. 
\citet{metarl_self_supervised} illustrates that adding contrastive objective to Varibad can improves the generalization and performance. 
\citet{decoupling} learns different meta-policies for exploration and exploitation to sidestep the issue of learning from informative trajectories with low rewards. 
Hyper-networks which generate the parameters of another network, can be more effective than conditioning the RL agent on task representations \citet{hypernetworks, hyper_recurrent}. 

The aim of offline meta-RL is to learn meta-policies from offline static datasets without further interaction. 
Contrastive objective for context encoder \cite{focal, corro} instead of RL objective improves the final performance and task inference. 
Context distribution shift can hinders the generalization ability to new tasks, especially when there is no prior context available. 
To alleviate this issue, CSRO \cite{csro} includes mutual information objective in the training of the context encoder, 
IDAC \cite{idac} filter context trajectories based on uncertainty, 
UNICORN \cite{unicorn} utilize a decoder to predict next states and observation based on task representation. 
GENTLE \cite{gentle} trains world models for each tasks and performs data augmentation for training the context encoder and the RL agent. 
Generative modeling can be applied to OMRL to sidestep bootstrapping.
\citet{prompting_dt} extends decision transformer \cite{dt} to OMRL setting by adding a short history of expert transitions (prompt) to the transformer model, adding the prompt significantly enhances the performance. 
Diffusion models \cite{meta_dd, diffusion_plan} has been used as a planner and data synthesizer for OMRL. 
Meta decision transformer \cite{meta_dt} trains context encoder based on the predictions of the next state and the reward, includes task representations in the trajectory and selects
prompts with highest prediction error.

\subsection{Representation Learning in RL}
Representation learning plays a critical role in RL where efficient state abstractions enable robust decision-making and policy optimization. 
In model-based RL, Dreamer \cite{dreamer1, dreamer2, dreamer3} learn abstract compact latent representations by utilizing auto-encoders and imagine transitions in the latent space to facilitate training. 
TD-MPC \cite{tdmpc, tdmpc2} and MuZero \cite{muzero} leverages latent world models to predict future outcomes and select best actions accordingly. 
In model-free approaches, TCRL \cite{tcrl}, TD7 \cite{td7} and IQRL \cite{iqrl} use self-supervised learning based on temporal consistency to learn representation.
For visual RL, SAC-AE \cite{sac_ae} enhances sample efficiency by learning latent representations directly from high-dimensional observations using auto-encoders. 
TED \cite{ted} and CMID \cite{cmid} improve generalization and stability by learning disentangled latent representation. 
TACO \cite{taco} learns abstract latent state and action spaces while utilizing contrastive objective. 

\section{Implementation Details \label{sec:implementation_detail}}

\begin{figure*}[hb]
    \centering
    \includegraphics[width=\linewidth]{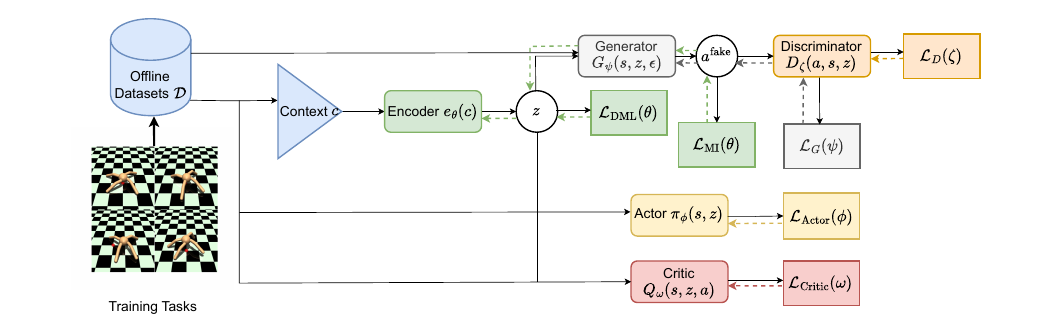}
    \caption{Overview of \our. 
    Solid black line illustrates forward computations and dashed lines illustrate the gradient paths in the computational graph. 
    The encoder is trained to infer the task from context based on distance metrics and mutual information objectives.}
    \label{fig:overview}
\end{figure*}

\cref{fig:overview} illustrate the procedure of training in \our. 
The context encoder is trained based on the distance metric objective to preserve the distance in embedding space and to minimize the mutual information objective, which reduces the impact of data collection on task representations. 
For the mutual information objective, we train a generator to sample in-distribution actions according to the datasets. 
Subsequently, we maximize the entropy of samples to achieve the mutual information objective. 
However, we require a discriminator for training the generator which is trained to distinguish between real and fake actions. 
The generator is trained to fool the discriminator. 
In summary, we trained a GAN to sample actions resembling the meta-behavior policy and maximize the entropy of generated samples to alleviate the context shift. 

We implemented \our with PyTorch \citep{pytorch} and used the Adam optimizer \cite{adam} for training the models. All components (context encoder, generator, discriminator, actor, and critic) are implemented as MLPs. 

\paragraph{Hardware} We used Nvidia A100, Nvidia V100, and Nvidia P100 GPUs to run our experiments. 
All experiments have been run on a single GPU with 2 number of CPU workers.

\subsection{Hyper-parameters \label{sec:hyperparams}}

\cref{tab:hyperparameters} lists all of the hyper-parameters for training \our which were used for the main experiments.
We find that the generator and discriminator architecture and learning rate are the most important hyper-parameters when training the GAN. 
We tuned these hyper-parameters on the Cheetah-Vel environment for 20K gradient steps, and used the same values across other environments. 
It is worth noting that the generated samples are not utilized directly for decision making but rather for regularization. 

\begin{table*}[htb]
\label{tab:hyperparameters}
\begin{center}
\begin{sc}
\begin{footnotesize}
\begin{tabular}{lll}
\toprule
Hyperparameter & Value & Description \\
\midrule
\textbf{Data Collection} & & \\
Train steps & $10^6$ & \\
Random steps & $5 \times 10^4$ &  Num. random steps at start \\
Num. eval episodes & $50$ & trajectories in offline datasets \\ 
Eval. every steps & $5\times 10^4$ & \\
Actor MLP dims & $[1024, 1024]$ & \\
Critic MLP dims & $[1024, 1024]$ & \\
Learning rate & $10^{-4}$ & actor, critic, entropy coeff learning rate \\
Batch size & $1024$ & \\
Discount factor $\gamma$ & $0.99$ & \\
Momentum coef. ($\tau$) & $0.005$ & soft update target network \\

\hline
\textbf{Offline Meta-RL} & & \\
Meta batch size & $16$ ($10$ ant-dir) & \\
Batch size & $256$ & \\
Context size & $256$ & \\
 & $512$ (ant-dir) & \\
 & $100$ (cheetah-vel) & \\
Buffer size & $2\times 10^5$ &  for each task\\
Discount factor $\gamma$ & $0.99$ & \\
Learning rate & $3 \times 10^{-4}$ & actor, critic and encoder\\
Actor/Critic MLP dims & $[256, 256, 256]$ &  \\
Encoder MLP dims & $[200, 200, 200]$ &  \\
Momentum coef. ($\tau$) & $0.005$ & soft update target network \\ 
Latent dim & $20$ ($40$ hopper and walker)& \\
behavior regularization & $50$ & \\
\hline
\textbf{mir-gan} & & \\
Generator MLP dims & $[200, 200, 200]$ &  \\
Discriminator MLP dims & $[256, 256]$ &  \\
Learning rate & $3 \times 10^{-4}$ & generator and discriminator\\
Num. update gan & $5$ & \\
Generator noise dim & $20$ &  \\
$\lambda$ coeff & $0.5$ &  \\
 & $0.25$ (humanoid-dir, cheetah-vel)&  \\
 \bottomrule
\end{tabular}
\end{footnotesize}
\end{sc}
\end{center}
\caption{Hyper-parameters of our method \our.}
\label{tab:hyperparameters}
\end{table*}

\subsection{Environments \label{sec:env}}

We evaluate \our in 8 environments from the MuJoCo \cite{mujoco} benchmark. 
In 4 of them, the reward function varies in each task, and in other ones, transition dynamics is different for each task. 
\cref{tab:envs} provides details of the environments we used, including the dimensionality of the observation and action space, and the distribution of train and test tasks.

\begin{table*}[htb]

\label{tab:envs}
\begin{center}
\begin{sc}
\begin{footnotesize}

\begin{tabular}{lcccc}
\toprule
\textbf{Environment} & \textbf{Observation dim} & \textbf{Action dim}  & \textbf{ID-Tasks}  & \textbf{OOD Tasks}\\
\midrule
cheetah-vel & 20 & 6 & $v \sim [1, 2]$ & $v \sim [0.5, 1] \cup [2, 2.5]$\\
ant-goal & 29 & 8 & $r=\{ 0.8, 1.2 \}$ & $r=1.6$\\
 & & & $\phi \sim [-\frac{\pi}{2}, \frac{\pi}{2}]$ & $\phi \sim [-\frac{\pi}{2}, \frac{\pi}{2}]$ \\
ant-dir & 29 & 8 & $\theta \sim [-\frac{\pi}{2}, \frac{\pi}{2}]$ & $\theta \sim [-\frac{3\pi}{4}, -\frac{\pi}{2}] \cup  [\frac{\pi}{2}, \frac{3\pi}{4}]$ \\
humanoid-dir & 376 & 17 & $\theta \sim [-\frac{\pi}{2}, \frac{\pi}{2}]$ & $\theta \sim [-\frac{3\pi}{4}, -\frac{\pi}{2}] \cup  [\frac{\pi}{2}, \frac{3\pi}{4}]$  \\
hopper-mass & 11 & 3 & $f_{\text{mass}} \sim [1.5^{-1}, 1.5^{1}]$ & $f_{\text{mass}} \sim [1.5^{-0.5}, 1.5^{-1}] \cup [1.5^{-1}, 1.5^{1.5} ]$   \\
hopper-friction & 11 & 3 & $f_{\text{friction}} \sim [1.5^{-1}, 1.5^{1}]$ & $f_{\text{friction}} \sim [1.5^{-0.5}, 1.5^{-1}] \cup [1.5^{-1}, 1.5^{1.5} ]$  \\
walker-mass & 17 & 6 & $f_{\text{mass}} \sim [1.5^{-1}, 1.5^{1}]$ & $f_{\text{mass}} \sim [1.5^{-0.5}, 1.5^{-1}] \cup [1.5^{-1}, 1.5^{1.5} ]$ \\
walker-friction & 17 & 6 & $f_{\text{friction}} \sim [1.5^{-1}, 1.5^{1}]$ & $f_{\text{friction}} \sim [1.5^{-0.5}, 1.5^{-1}] \cup [1.5^{-1}, 1.5^{1.5}] $  \\
\bottomrule
\end{tabular}
\end{footnotesize}
\end{sc}
\end{center}
\caption{Environment used for evaluation of different methods.}
\label{tab:envs}
\end{table*}

\section{Further Experiments \label{sec:extra_exp}}

\subsection{Non-prior Context \label{sec:nonprior}}
In testing, the agent collects the context $c$ by interacting with the environment and infers the task by embedding context to task representations according to \cref{eq:encoder}. 
In online context collection, the learned agent is conditioned on prior task representations $z_0$ to collect the transitions in context, update the task representation, and collect more transitions according to the new task representation. This process is iterated until there are enough transitions in the context. 
However, the distribution of context $c$ will correlate with the prior task representations $z_0$ since the learned agent is used to collect context in different tasks. 

\cite{csro} proposed non-prior context collection where the agent initially explores the task independently and randomly for several steps. 
Then the agent infers the task by computing task representations according to \cref{eq:encoder} and then is conditioned on the task representations and collects more transitions. 
Similar to online testing, the process of collecting new transitions and updating the task representations is iterated until there are enough transitions in the context. 

\cref{tab:nonprior} compare the performance of \our with the baselines for both in-distribution and out-of-distribution tasks. 
Similar to offline and online context collections, \our outperforms the baselines in most of the environments and generalizes better to out-of-distribution tasks. 

\begin{table*} [htb]
\centering
\begin{tabular}{c|c|ccccc} 
\hline\hline
\multicolumn{1}{l}{\sc Environment} & \multicolumn{1}{l}{\sc Task}  & {\sc offlinepearl} & {\sc focal} & \sc{csro} & {\sc unicorn} & \our (Ours) \\ 
\hline
 Cheetah-Vel     & \multirow{8}{*}{ID}      & $ 74.82 \pm 17.06 $ & $ 89.11 \pm 1.60 $  & $ 91.69 \pm 1.04 $  & $ 91.11 \pm 0.81 $  & $ 94.21 \pm 0.35 $  \\
 Ant-Goal        &                          & $ 57.07 \pm 2.78 $  & $ 90.09 \pm 3.50 $  & $ 94.24 \pm 3.73 $  & $ 91.71 \pm 2.50 $  & $ 102.18 \pm 2.68 $ \\
 Ant-Dir         &                          & $ 30.61 \pm 5.20 $  & $ 62.84 \pm 2.69 $  & $ 65.10 \pm 4.17 $  & $ 59.77 \pm 6.23 $  & $ 62.34 \pm 12.69 $ \\
 Humanoid-Dir    &                          & $ 66.72 \pm 3.91 $  & $ 43.62 \pm 6.72 $  & $ 51.02 \pm 6.84 $  & $ 43.70 \pm 3.79 $  & $ 60.72 \pm 5.41 $  \\
 Hopper-Mass     &                          & $ 85.87 \pm 9.87 $  & $ 84.07 \pm 9.69 $  & $ 88.69 \pm 4.07 $  & $ 88.14 \pm 7.22 $  & $ 100.94 \pm 1.55 $ \\
 Hopper-Friction &                          & $ 66.76 \pm 10.19 $ & $ 81.23 \pm 15.70 $ & $ 82.78 \pm 14.25 $ & $ 81.05 \pm 11.39 $ & $ 91.02 \pm 5.47 $  \\
 Walker-Mass     &                          & $ 37.15 \pm 10.76 $ & $ 37.50 \pm 6.63 $  & $ 46.59 \pm 4.62 $  & $ 38.67 \pm 8.32 $  & $ 68.64 \pm 3.36 $  \\
 Walker-Friction &                          & $ 37.86 \pm 14.29 $ & $ 33.63 \pm 6.93 $  & $ 40.55 \pm 11.38 $ & $ 35.35 \pm 7.76 $  & $ 49.85 \pm 8.32 $  \\
\hline
 Cheetah-Vel     & \multirow{8}{*}{OOD}     & $ 55.36 \pm 5.59 $  & $ 66.88 \pm 3.38 $  & $ 63.68 \pm 3.28 $  & $ 62.08 \pm 1.75 $  & $ 70.38 \pm 1.22 $ \\
 Ant-Goal        &                          & $ 23.63 \pm 4.12 $  & $ 48.26 \pm 3.06 $  & $ 44.78 \pm 4.21 $  & $ 47.61 \pm 4.00 $  & $ 67.26 \pm 5.82 $ \\
 Ant-Dir         &                          & $ -22.99 \pm 4.88 $ & $ 23.67 \pm 3.21 $  & $ 16.76 \pm 4.20 $  & $ 23.57 \pm 4.39 $  & $ 32.31 \pm 5.51 $ \\
 Humanoid-Dir    &                          & $ 61.08 \pm 6.33 $  & $ 37.45 \pm 5.70 $  & $ 43.52 \pm 8.90 $  & $ 34.64 \pm 4.20 $  & $ 51.12 \pm 3.25 $ \\
 Hopper-Mass     &                          & $ 73.63 \pm 8.35 $  & $ 79.07 \pm 10.80 $ & $ 84.70 \pm 9.14 $  & $ 86.67 \pm 8.75 $  & $ 96.60 \pm 2.73 $ \\
 Hopper-Friction &                          & $ 49.23 \pm 8.33 $  & $ 56.08 \pm 9.58 $  & $ 62.00 \pm 8.04 $  & $ 55.20 \pm 10.30 $ & $ 65.74 \pm 5.20 $ \\
 Walker-Mass     &                          & $ 34.99 \pm 5.39 $  & $ 39.68 \pm 9.21 $  & $ 42.28 \pm 3.94 $  & $ 46.06 \pm 8.60 $  & $ 63.89 \pm 3.30 $ \\
 Walker-Friction &                          & $ 42.29 \pm 16.20 $ & $ 36.50 \pm 6.22 $  & $ 42.29 \pm 12.45 $ & $ 40.86 \pm 9.03 $  & $ 54.75 \pm 8.14 $ \\
\hline\hline
\end{tabular}
\caption{The average normalized return for non-prior context collection strategy after 100k training steps, averaged over 5 random seeds, $\pm$ represents standard deviation.}
\label{tab:nonprior}
\end{table*}

\subsection{t-SNE visualizations \label{sec:tsne}}
\begin{figure*}
    \centering
    \includegraphics[width=\linewidth]{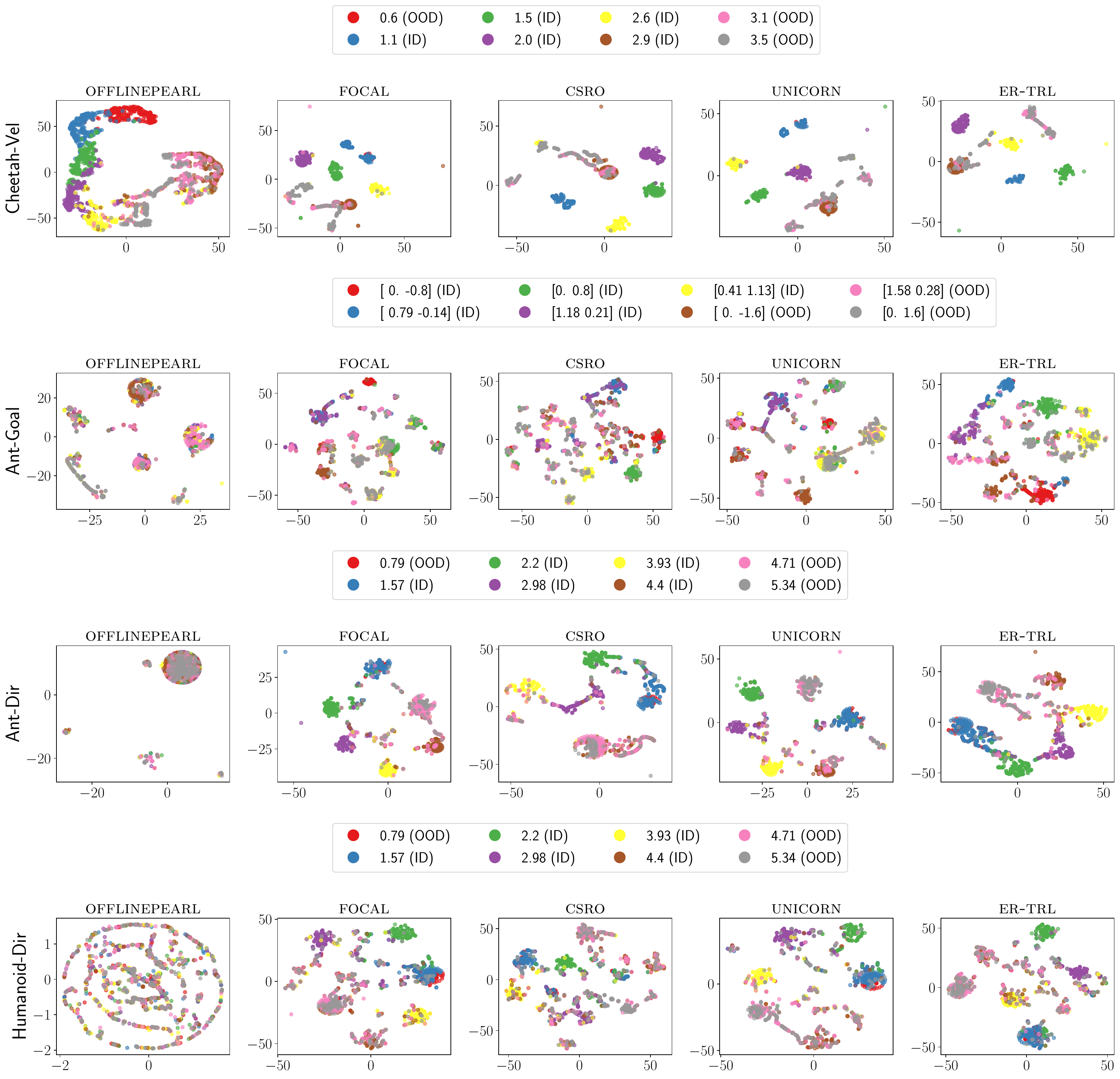}
    \caption{t-SNE visualization for reward changing environments.}
    \label{fig:tsne_reward}
\end{figure*}

\begin{figure*}
    \centering
    \includegraphics[width=\linewidth]{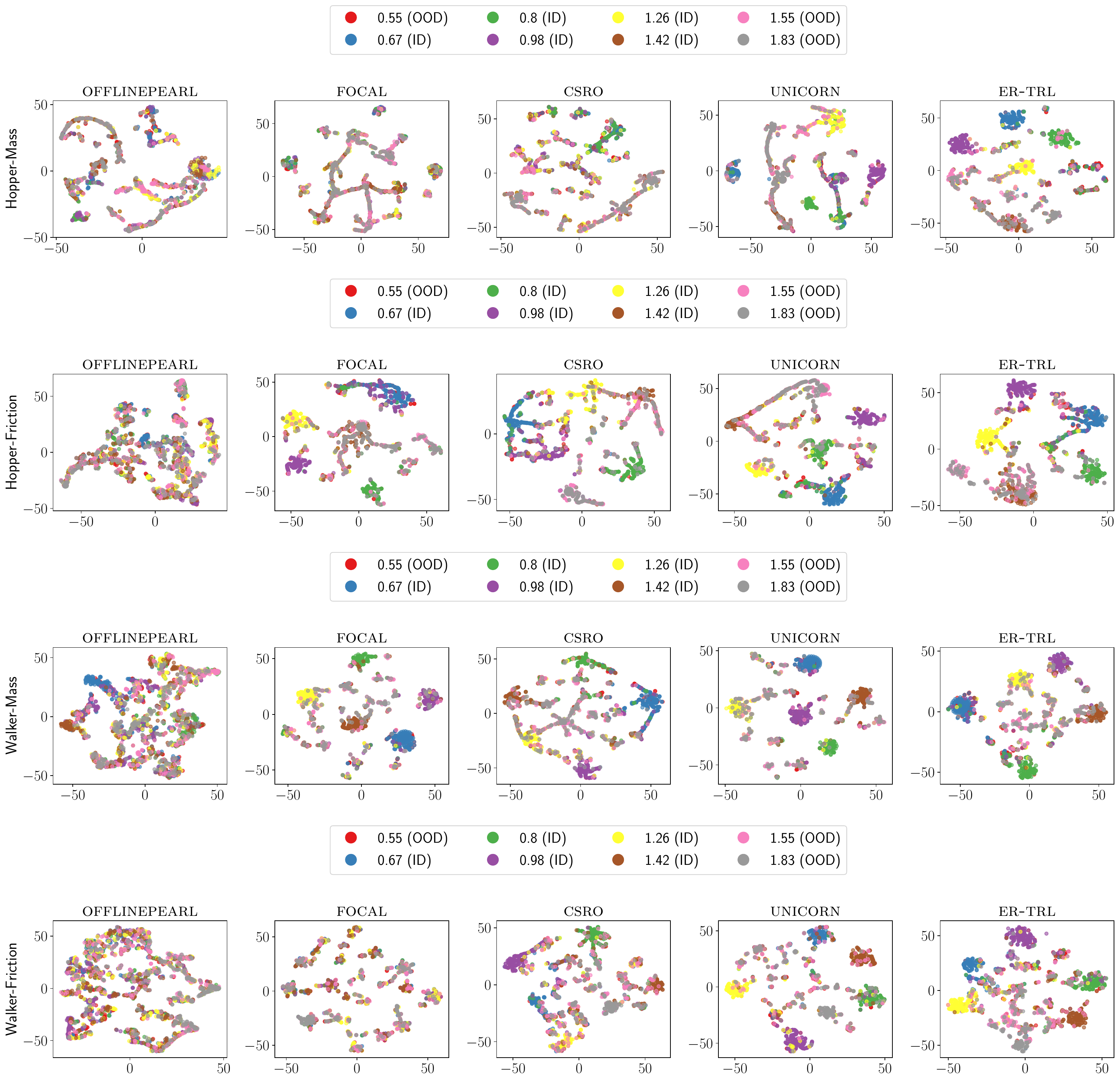}
    \caption{t-SNE visualization for dynamic changing environment.}
    \label{fig:tsne_dynamics}
\end{figure*}

We provide further visualization of task representations in \cref{fig:tsne_reward} and \cref{fig:tsne_dynamics} by projecting them in 2d space using t-SNE \cite{tsne}. 
OfflinePearl is unable to learn distinguishable task representations while other methods that use distance metric objective learn better representations in most environments. 
OfflinePearl learns reasonable task representations only in the Cheetah-Vel environment. 
\our learns distinct task representations for each task in most of the environments while trying to preserve the space in embedding space. 
\eg in Hopper-Mass and Hopper-Friction \our learns better task representations compared to others; each task can be separated more easily while distance is preserved.  

\subsection{Ablation: Distance Metric Learning \label{sec:ablation}}
\begin{figure*}
    \centering
    \includegraphics[width=\linewidth]{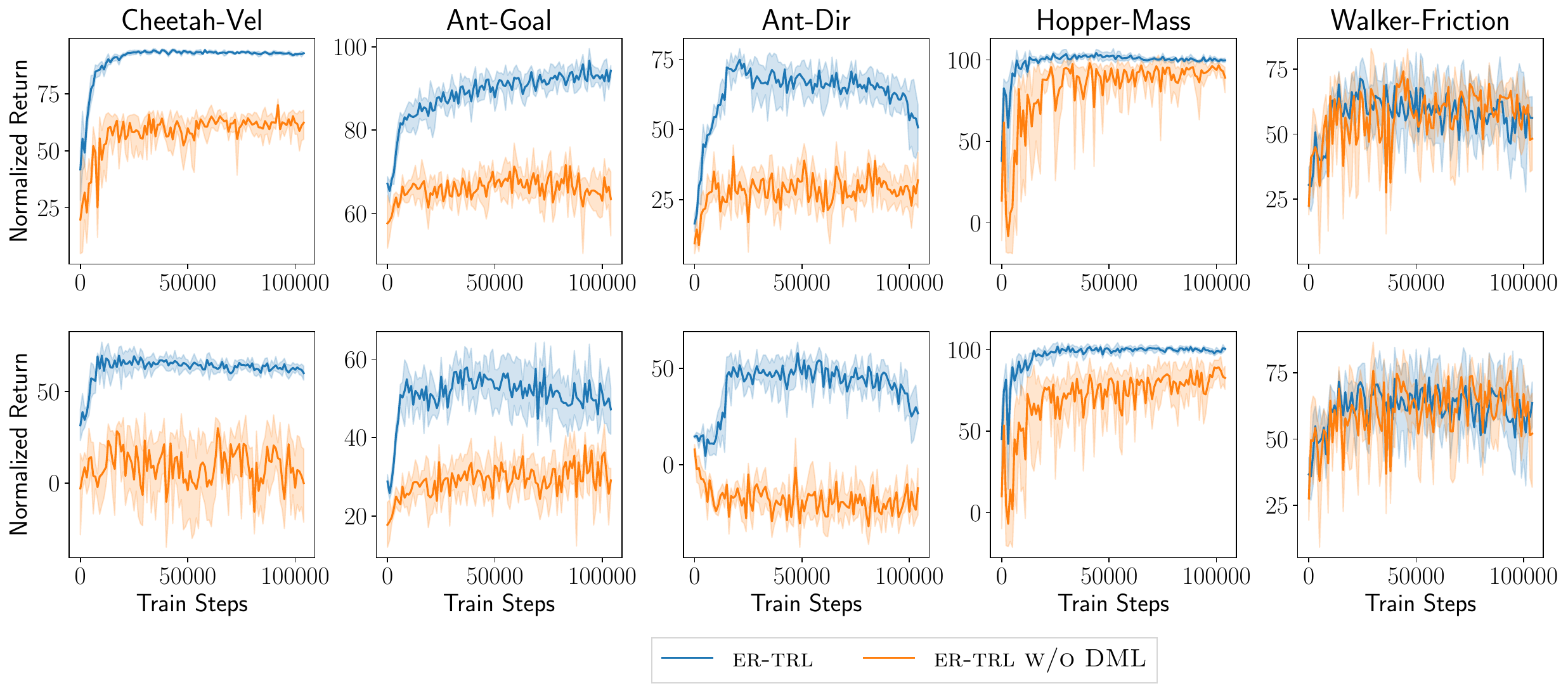}
    \caption{The importance of distance metric learning in training context encoder. 
    The top and bottom rows illustrate the adaptation performance for ID and OOD tasks during training. 
    We plot the mean (solid line) and the 95\% confidence intervals (shaded) across 5 random seeds, where each seed averages over 10 evaluation episodes.}
    \label{fig:ablation}
\end{figure*}

In this section, we evaluate the necessity of the distance metric objective for training the context encoder. 
As stated, the distance metric objective encourages the context encoder to learn distinct task representations by minimizing the distance between transitions from the same task in embedding space and maximizing the distance between transitions from different tasks. 
\cref{fig:ablation} demonstrate the necessity of this objective, when we ignore this objective, there is a sharp drop in the performance for both in-distribution and out-of-distribution tasks. 
This drop in performance is more significant for reward-changing environments. 

\subsection{Performance Plots \label{sec:plots}}
\begin{figure*}
    \centering
    \includegraphics[width=\linewidth]{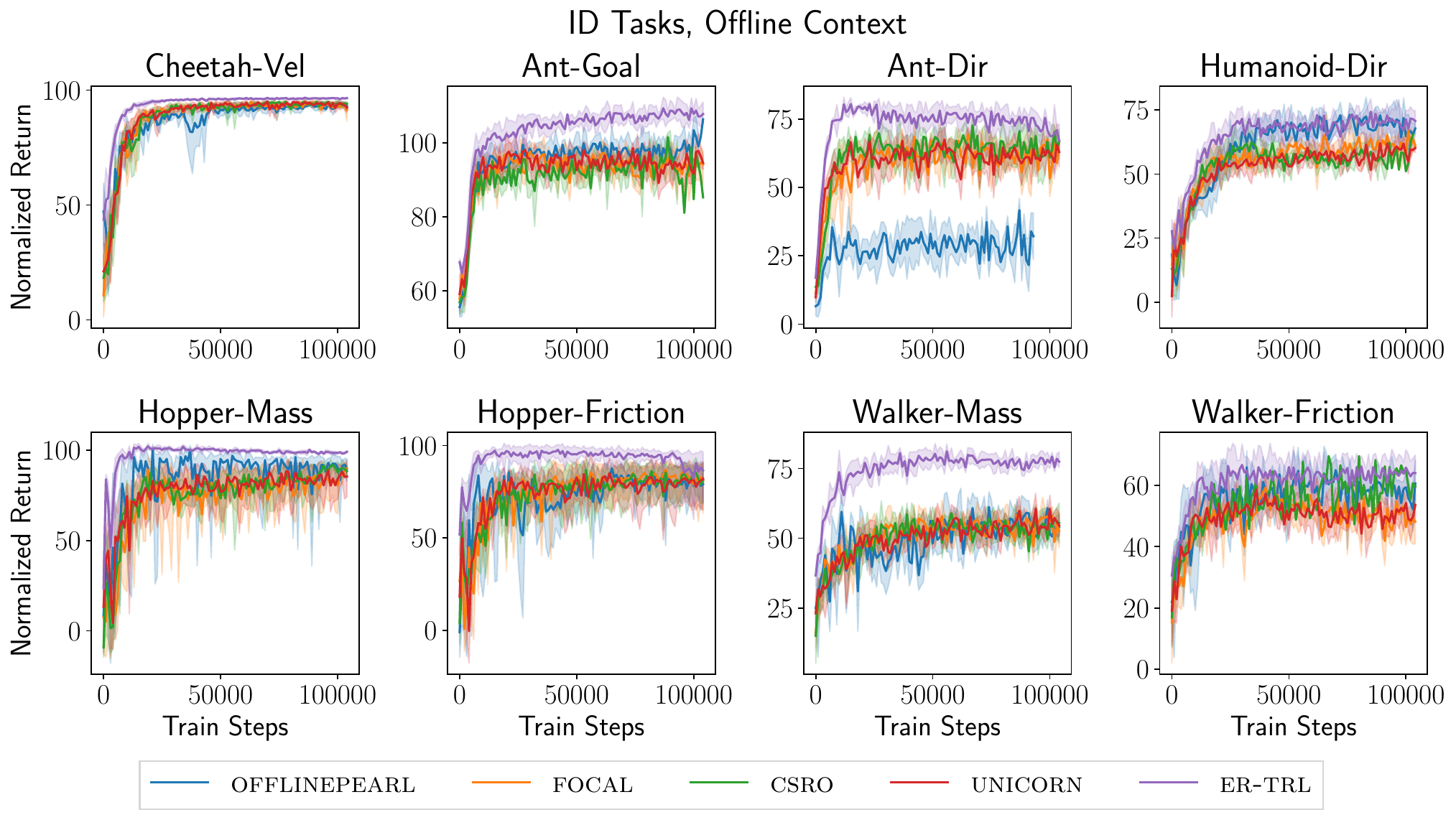}
    \caption{Learning curve of different methods for ID tasks and offline context. 
    We plot the mean (solid line) and the 95\% confidence intervals (shaded) across 5 random seeds, where each seed averages over 10 evaluation episodes.}
    \label{fig:plot_id_offline}
\end{figure*}

\begin{figure*}
    \centering
    \includegraphics[width=\linewidth]{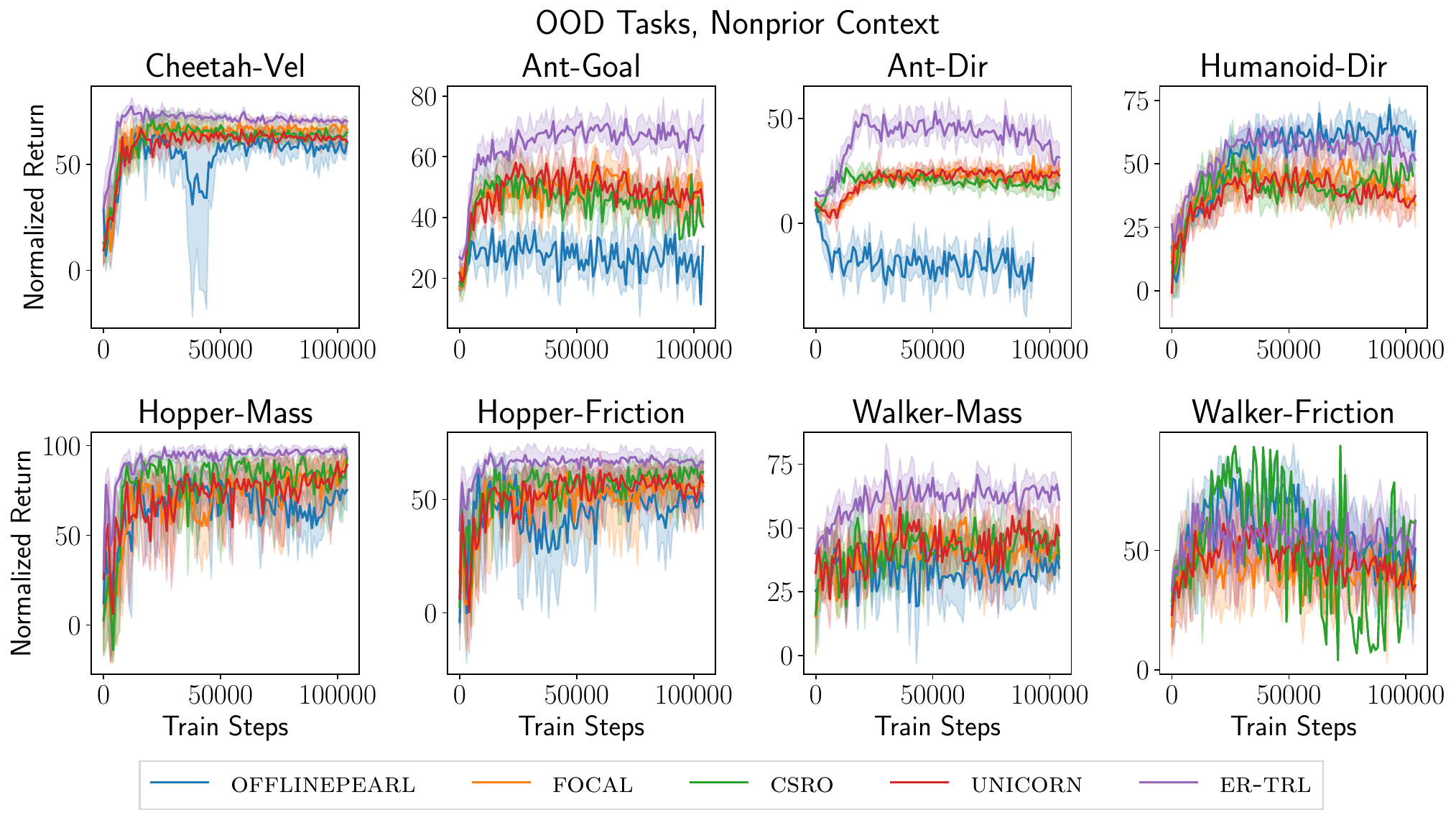}
    \caption{Learning curve of different methods for OOD tasks and non-prior context. 
    We plot the mean (solid line) and the 95\% confidence intervals (shaded) across 5 random seeds, where each seed averages over 10 evaluation episodes.}
    \label{fig:plot_ood_nonprior}
\end{figure*}

In this section, we provide learning curves (\cref{fig:plot_id_offline} till \cref{fig:plot_ood_nonprior}) of different methods during training for both in-distribution tasks and out-of-distribution tasks with different context collection strategies. 
\our outperforms the baselines in most of the experiments, specially in out-of-distribution tasks with online and non-prior context collection strategies.

\begin{figure*}
    \centering
    \includegraphics[width=\linewidth]{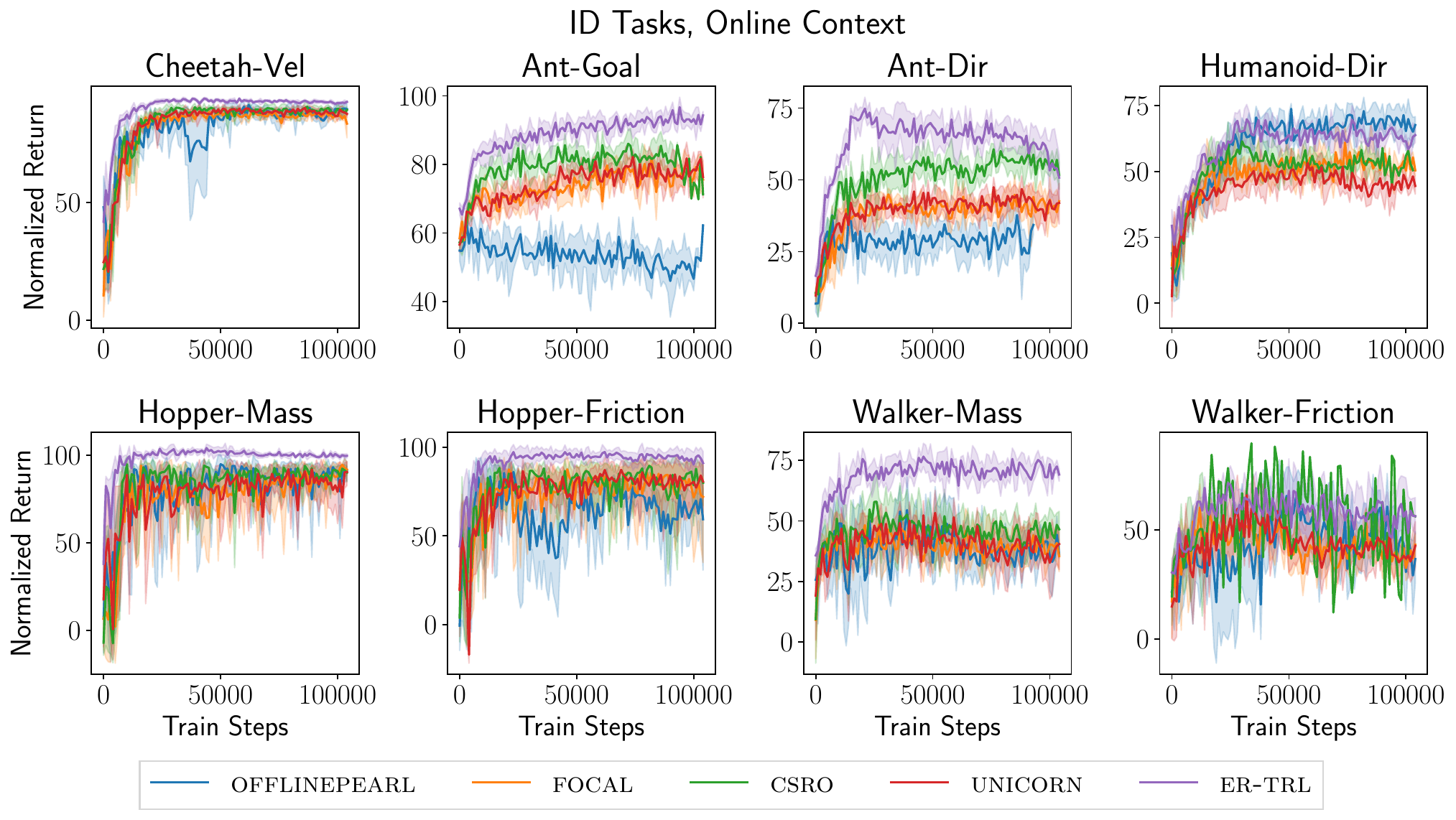}
    \caption{Learning curve of different methods for ID tasks and online context. 
    We plot the mean (solid line) and the 95\% confidence intervals (shaded) across 5 random seeds, where each seed averages over 10 evaluation episodes.}
    \label{fig:plot_id_online}
\end{figure*}

\begin{figure*}
    \centering
    \includegraphics[width=\linewidth]{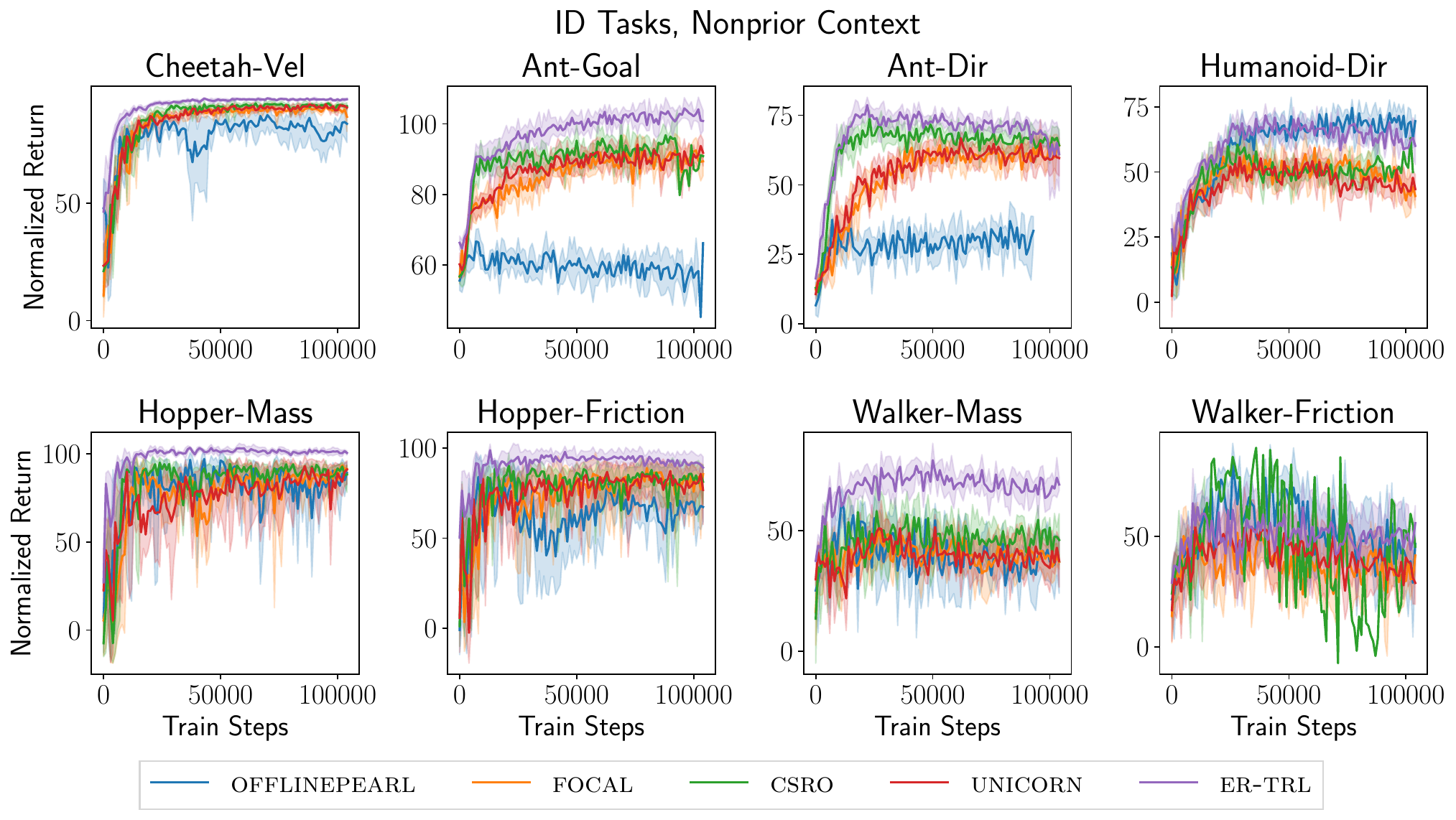}
    \caption{Learning curve of different methods for ID tasks and non-prior context. We plot the mean (solid line) and the 95\% confidence intervals (shaded) across 5 random seeds, where each seed averages over 10 evaluation episodes.}
    \label{fig:plot_id_nonprior}
\end{figure*}

\begin{figure*}
    \centering
    \includegraphics[width=\linewidth]{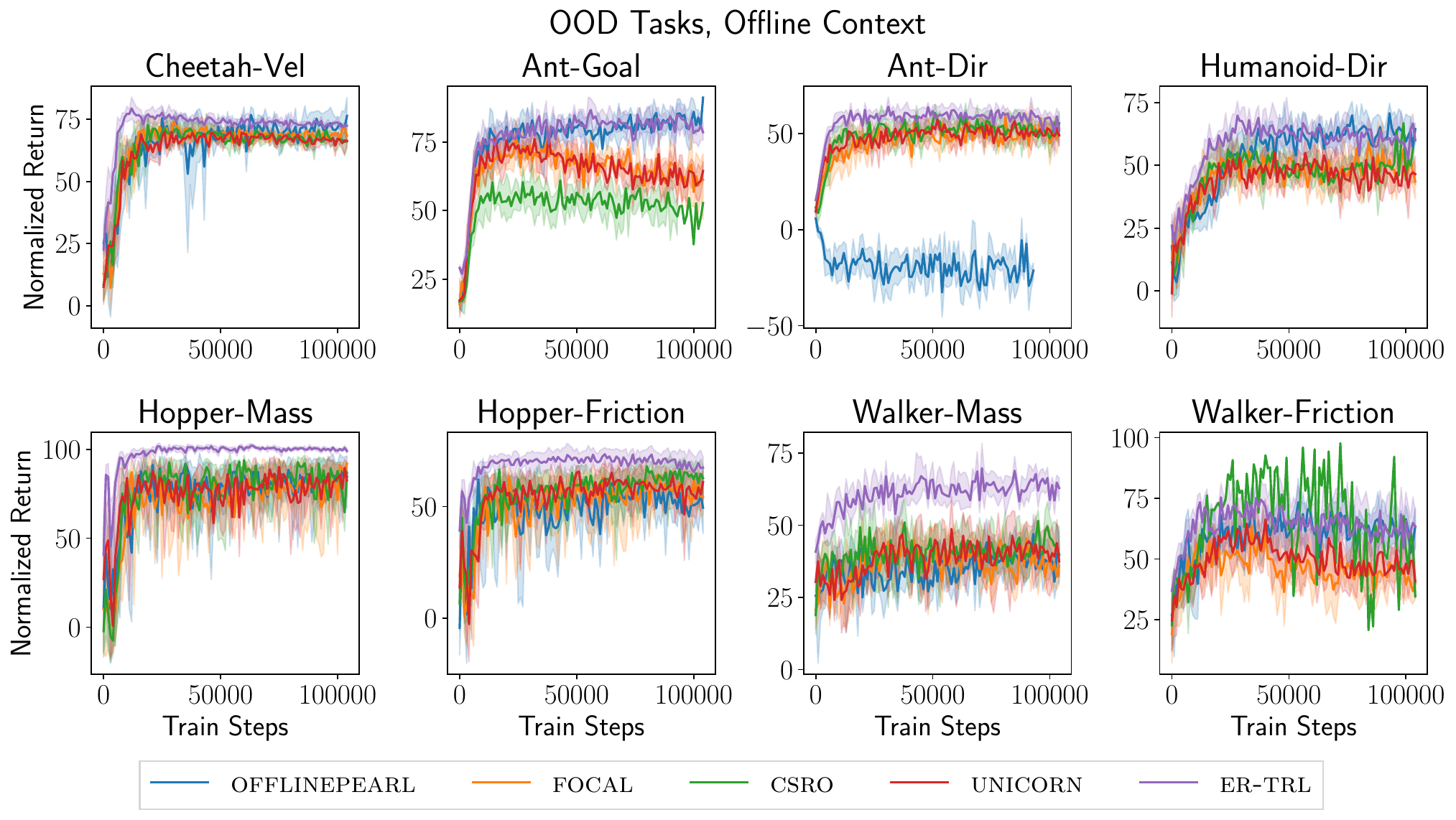}
    \caption{Learning curve of different methods for OOD tasks and offline context. 
    We plot the mean (solid line) and the 95\% confidence intervals (shaded) across 5 random seeds, where each seed averages over 10 evaluation episodes.}
    \label{fig:plot_ood_offline}
\end{figure*}

\begin{figure*}
    \centering
    \includegraphics[width=\linewidth]{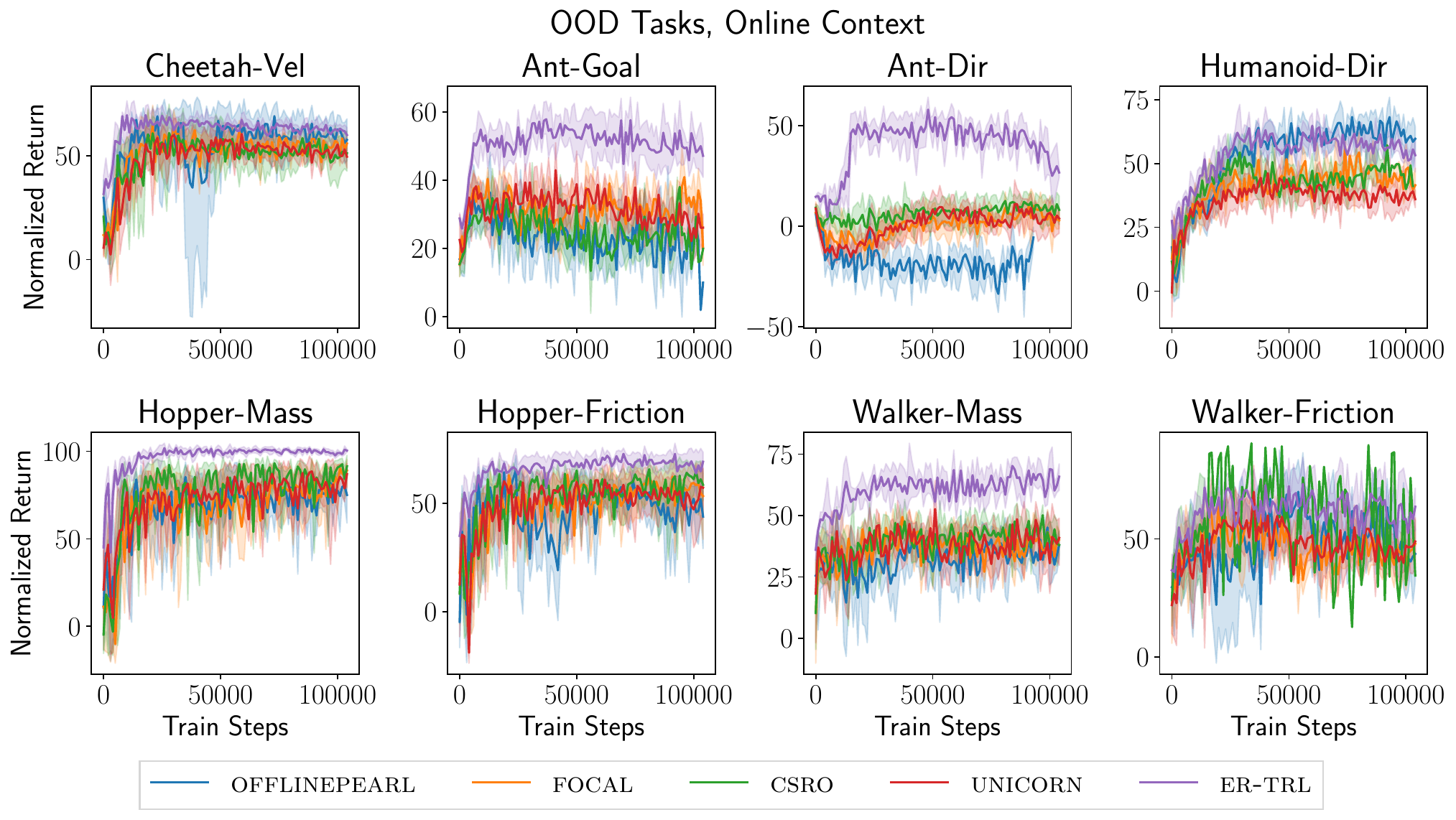}
    \caption{Learning curve of different methods for OOD tasks and online context. 
    We plot the mean (solid line) and the 95\% confidence intervals (shaded) across 5 random seeds, where each seed averages over 10 evaluation episodes.}
    \label{fig:plot_ood_online}
\end{figure*}

\subsection{Unnormalized Returns \label{sec:unnormalized}}
We provide unnormalized returns for a fair comparison to previous implementations. 
\cref{tab:id_all} summarize the results of different methods for in-distribution tasks with different context collection strategies.
\cref{tab:ood_all} similarly illustrates the results for out-of-distribution tasks. 
\begin{table*} [h]
\centering
\begin{tabular}{c|c|ccccc} 
\hline\hline
\multicolumn{1}{l}{\sc Environment} & \multicolumn{1}{l}{\sc Context}  & {\sc offlinepearl} & {\sc focal} & \sc{csro} & {\sc unicorn} & \our (Ours) \\ 
\hline
 Cheetah-Vel     & \multirow{8}{*}{Offline} & $ -88.7 \pm 86.1 $  & $ -48.9 \pm 4.4 $  & $ -45.2 \pm 3.1 $   & $ -46.5 \pm 3.3 $  & $ -35.5 \pm 0.8 $  \\
 Ant-Goal        &                          & $ -160.5 \pm 10.6 $ & $ -176.0 \pm 8.5 $ & $ -173.5 \pm 10.2 $ & $ -171.4 \pm 8.4 $ & $ -134.2 \pm 7.5 $ \\
 Ant-Dir         &                          & $ 322.8 \pm 26.6 $  & $ 609.3 \pm 14.6 $ & $ 647.2 \pm 40.2 $  & $ 624.6 \pm 17.6 $ & $ 680.9 \pm 39.7 $ \\
 Humanoid-Dir    &                          & $ 707.8 \pm 36.1 $  & $ 671.5 \pm 17.6 $ & $ 621.3 \pm 11.7 $  & $ 636.7 \pm 19.4 $ & $ 745.5 \pm 48.4 $ \\
 Hopper-Mass     &                          & $ 569.7 \pm 47.7 $  & $ 521.0 \pm 53.2 $ & $ 540.7 \pm 59.9 $  & $ 553.0 \pm 45.8 $ & $ 614.7 \pm 1.6 $  \\
 Hopper-Friction &                          & $ 511.7 \pm 56.7 $  & $ 517.5 \pm 71.5 $ & $ 526.0 \pm 48.5 $  & $ 515.7 \pm 65.7 $ & $ 546.0 \pm 74.4 $ \\
 Walker-Mass     &                          & $ 449.4 \pm 16.2 $  & $ 456.3 \pm 33.1 $ & $ 464.0 \pm 8.9 $   & $ 473.9 \pm 14.4 $ & $ 618.8 \pm 10.7 $ \\
 Walker-Friction &                          & $ 490.4 \pm 30.1 $  & $ 432.3 \pm 25.2 $ & $ 439.0 \pm 40.7 $  & $ 453.5 \pm 15.1 $ & $ 541.3 \pm 47.8 $ \\

\hline
 Cheetah-Vel     & \multirow{8}{*}{Online} & $ -100.9 \pm 81.4 $ & $ -76.0 \pm 7.0 $   & $ -62.9 \pm 7.1 $   & $ -67.3 \pm 7.9 $  & $ -50.7 \pm 2.0 $  \\
 Ant-Goal        &                          & $ -314.4 \pm 9.8 $  & $ -225.1 \pm 9.4 $  & $ -214.9 \pm 12.6 $ & $ -219.9 \pm 6.4 $ & $ -177.3 \pm 6.2 $ \\
 Ant-Dir         &                          & $ 328.3 \pm 33.5 $  & $ 443.0 \pm 41.7 $  & $ 544.5 \pm 32.5 $  & $ 433.8 \pm 42.7 $ & $ 535.4 \pm 97.2 $ \\
 Humanoid-Dir    &                          & $ 713.1 \pm 49.8 $  & $ 604.5 \pm 17.1 $  & $ 584.9 \pm 26.0 $  & $ 522.2 \pm 18.6 $ & $ 665.4 \pm 21.0 $ \\
 Hopper-Mass     &                          & $ 562.5 \pm 42.7 $  & $ 554.5 \pm 40.0 $  & $ 565.7 \pm 17.0 $  & $ 545.0 \pm 59.0 $ & $ 618.7 \pm 5.3 $  \\
 Hopper-Friction &                          & $ 446.4 \pm 65.0 $  & $ 487.1 \pm 144.8 $ & $ 523.5 \pm 71.8 $  & $ 516.8 \pm 66.8 $ & $ 575.7 \pm 31.0 $ \\
 Walker-Mass     &                          & $ 338.6 \pm 60.7 $  & $ 355.6 \pm 31.5 $  & $ 404.2 \pm 60.8 $  & $ 341.4 \pm 69.8 $ & $ 569.7 \pm 14.7 $ \\
 Walker-Friction &                          & $ 353.0 \pm 58.9 $  & $ 371.7 \pm 35.9 $  & $ 419.0 \pm 57.6 $  & $ 368.7 \pm 44.2 $ & $ 473.9 \pm 71.8 $ \\
\hline
 Cheetah-Vel     & \multirow{8}{*}{Non-prior} & $ -125.2 \pm 71.8 $ & $ -65.2 \pm 6.2 $   & $ -55.1 \pm 4.9 $   & $ -56.9 \pm 3.2 $  & $ -44.4 \pm 1.8 $   \\
 Ant-Goal        &                          & $ -287.2 \pm 8.0 $  & $ -187.4 \pm 10.2 $ & $ -174.2 \pm 11.3 $ & $ -182.9 \pm 8.1 $ & $ -150.0 \pm 8.0 $  \\
 Ant-Dir         &                          & $ 341.8 \pm 44.6 $  & $ 614.7 \pm 23.2 $  & $ 638.0 \pm 34.7 $  & $ 588.8 \pm 53.9 $ & $ 612.3 \pm 107.0 $ \\
 Humanoid-Dir    &                          & $ 711.1 \pm 34.4 $  & $ 511.1 \pm 58.2 $  & $ 575.8 \pm 59.4 $  & $ 511.5 \pm 32.9 $ & $ 660.8 \pm 47.3 $  \\
 Hopper-Mass     &                          & $ 551.8 \pm 47.6 $  & $ 542.7 \pm 47.7 $  & $ 565.7 \pm 19.2 $  & $ 561.7 \pm 36.3 $ & $ 624.6 \pm 7.8 $   \\
 Hopper-Friction &                          & $ 450.6 \pm 48.8 $  & $ 519.2 \pm 75.1 $  & $ 527.4 \pm 68.5 $  & $ 519.1 \pm 54.6 $ & $ 567.0 \pm 26.1 $  \\
 Walker-Mass     &                          & $ 341.8 \pm 72.9 $  & $ 344.0 \pm 44.1 $  & $ 407.4 \pm 30.9 $  & $ 352.3 \pm 57.3 $ & $ 556.0 \pm 23.1 $  \\
 Walker-Friction &                          & $ 362.5 \pm 90.3 $  & $ 336.4 \pm 45.7 $  & $ 376.9 \pm 71.9 $  & $ 348.1 \pm 51.6 $ & $ 438.6 \pm 52.1 $  \\
\hline\hline
\end{tabular}
\caption{The average return for in-distribution tasks after 100k training steps, averaged over 5 random seeds, $\pm$ represents standard deviation.}
\label{tab:id_all}
\end{table*}

\begin{table*}[h]
\centering

\begin{tabular}{c|c|ccccc} 

\hline\hline
\multicolumn{1}{l}{\sc Environment} & \multicolumn{1}{l}{\sc Context}  & {\sc offlinepearl} & {\sc focal} & \sc{csro} & {\sc unicorn} & \our (Ours) \\ 
\hline
 Cheetah-Vel     & \multirow{8}{*}{Offline} & $ -140.2 \pm 83.0 $ & $ -116.0 \pm 3.2 $  & $ -117.2 \pm 8.9 $  & $ -120.9 \pm 5.1 $ & $ -101.2 \pm 4.5 $  \\
 Ant-Goal        &                          & $ -249.8 \pm 13.9 $ & $ -309.3 \pm 16.5 $ & $ -334.8 \pm 11.3 $ & $ -308.4 \pm 9.8 $ & $ -258.0 \pm 11.4 $ \\
 Ant-Dir         &                          & $ -68.8 \pm 26.5 $  & $ 472.0 \pm 24.0 $  & $ 499.4 \pm 33.4 $  & $ 491.5 \pm 19.0 $ & $ 531.0 \pm 24.5 $  \\
 Humanoid-Dir    &                          & $ 702.5 \pm 60.1 $  & $ 596.3 \pm 22.6 $  & $ 595.4 \pm 24.5 $  & $ 561.3 \pm 19.3 $ & $ 691.4 \pm 24.1 $  \\
 Hopper-Mass     &                          & $ 544.6 \pm 67.7 $  & $ 525.3 \pm 65.2 $  & $ 524.9 \pm 30.2 $  & $ 541.6 \pm 33.9 $ & $ 611.4 \pm 3.7 $   \\
 Hopper-Friction &                          & $ 379.3 \pm 46.9 $  & $ 406.5 \pm 35.6 $  & $ 430.8 \pm 41.5 $  & $ 399.7 \pm 37.6 $ & $ 450.8 \pm 30.6 $  \\
 Walker-Mass     &                          & $ 323.1 \pm 14.7 $  & $ 314.3 \pm 33.9 $  & $ 359.3 \pm 28.3 $  & $ 335.6 \pm 56.9 $ & $ 473.0 \pm 10.9 $  \\
 Walker-Friction &                          & $ 462.8 \pm 33.0 $  & $ 334.0 \pm 42.4 $  & $ 398.9 \pm 35.0 $  & $ 374.4 \pm 44.8 $ & $ 480.8 \pm 69.8 $  \\
\hline
 Cheetah-Vel     & \multirow{8}{*}{Online}  & $ -154.1 \pm 72.3 $ & $ -131.9 \pm 9.4 $  & $ -136.5 \pm 22.8 $ & $ -137.5 \pm 7.6 $  & $ -109.3 \pm 2.4 $  \\
 Ant-Goal        &                          & $ -428.1 \pm 16.2 $ & $ -400.4 \pm 14.7 $ & $ -410.0 \pm 12.9 $ & $ -409.6 \pm 15.4 $ & $ -342.9 \pm 12.0 $ \\
 Ant-Dir         &                          & $ -80.5 \pm 26.7 $  & $ 112.9 \pm 32.7 $  & $ 148.3 \pm 56.2 $  & $ 118.9 \pm 66.7 $  & $ 303.6 \pm 44.0 $  \\
 Humanoid-Dir    &                          & $ 686.1 \pm 61.1 $  & $ 556.5 \pm 32.9 $  & $ 576.5 \pm 50.4 $  & $ 488.6 \pm 16.0 $  & $ 633.4 \pm 28.7 $  \\
 Hopper-Mass     &                          & $ 513.3 \pm 52.2 $  & $ 530.4 \pm 42.3 $  & $ 542.5 \pm 51.1 $  & $ 536.1 \pm 28.2 $  & $ 612.0 \pm 6.2 $   \\
 Hopper-Friction &                          & $ 361.8 \pm 51.8 $  & $ 400.5 \pm 59.3 $  & $ 419.5 \pm 42.0 $  & $ 384.7 \pm 73.1 $  & $ 445.5 \pm 38.5 $  \\
 Walker-Mass     &                          & $ 309.2 \pm 34.5 $  & $ 337.2 \pm 30.2 $  & $ 342.7 \pm 29.6 $  & $ 335.3 \pm 46.9 $  & $ 489.8 \pm 22.4 $  \\
 Walker-Friction &                          & $ 348.0 \pm 59.5 $  & $ 371.5 \pm 38.1 $  & $ 399.3 \pm 58.3 $  & $ 381.7 \pm 45.0 $  & $ 457.2 \pm 62.6 $  \\
\hline
 Cheetah-Vel     & \multirow{8}{*}{Non-prior} & $ -154.1 \pm 72.3 $ & $ -131.9 \pm 9.4 $  & $ -136.5 \pm 22.8 $ & $ -137.5 \pm 7.6 $  & $ -109.3 \pm 2.4 $  \\
 Ant-Goal        &                          & $ -428.1 \pm 16.2 $ & $ -400.4 \pm 14.7 $ & $ -410.0 \pm 12.9 $ & $ -409.6 \pm 15.4 $ & $ -342.9 \pm 12.0 $ \\
 Ant-Dir         &                          & $ -80.5 \pm 26.7 $  & $ 112.9 \pm 32.7 $  & $ 148.3 \pm 56.2 $  & $ 118.9 \pm 66.7 $  & $ 303.6 \pm 44.0 $  \\
 Humanoid-Dir    &                          & $ 686.1 \pm 61.1 $  & $ 556.5 \pm 32.9 $  & $ 576.5 \pm 50.4 $  & $ 488.6 \pm 16.0 $  & $ 633.4 \pm 28.7 $  \\
 Hopper-Mass     &                          & $ 513.3 \pm 52.2 $  & $ 530.4 \pm 42.3 $  & $ 542.5 \pm 51.1 $  & $ 536.1 \pm 28.2 $  & $ 612.0 \pm 6.2 $   \\
 Hopper-Friction &                          & $ 361.8 \pm 51.8 $  & $ 400.5 \pm 59.3 $  & $ 419.5 \pm 42.0 $  & $ 384.7 \pm 73.1 $  & $ 445.5 \pm 38.5 $  \\
 Walker-Mass     &                          & $ 309.2 \pm 34.5 $  & $ 337.2 \pm 30.2 $  & $ 342.7 \pm 29.6 $  & $ 335.3 \pm 46.9 $  & $ 489.8 \pm 22.4 $  \\
 Walker-Friction &                          & $ 348.0 \pm 59.5 $  & $ 371.5 \pm 38.1 $  & $ 399.3 \pm 58.3 $  & $ 381.7 \pm 45.0 $  & $ 457.2 \pm 62.6 $  \\
\hline\hline
\end{tabular}
\caption{The average return for out-of-distribution test tasks after 100k training steps, averaged over 5 random seeds, $\pm$ represents standard deviation.}
\label{tab:ood_all}

\end{table*}
\end{document}